\documentclass[10pt,twocolumn,letterpaper]{article}

\usepackage[pagenumbers]{wacv} 

\usepackage[utf8]{inputenc} 
\usepackage[T1]{fontenc}    
\usepackage{url}            
\usepackage{booktabs}       
\usepackage{amsfonts}       
\usepackage{nicefrac}       
\usepackage{microtype}      
\usepackage{xcolor}         
\usepackage{graphicx}
\usepackage{subcaption}
\usepackage{amsmath}
\usepackage{amssymb}
\usepackage{mathtools}
\usepackage{amsthm}
\usepackage{multirow}
\usepackage{soul}
\usepackage{makecell}

\newcommand{\dataset}{SynthGait-19K}
\newcommand{\pipeline}{Gait2Vid}
\newcommand{\model}{GaitXFormer}

\definecolor{wacvblue}{rgb}{0.21,0.49,0.74}
\usepackage[pagebackref,breaklinks,colorlinks,allcolors=wacvblue]{hyperref}
\usepackage[capitalize,noabbrev]{cleveref}

\def\wacvPaperID{3481} 
\def\confName{WACV}
\def\confYear{2027}

\title{SynthGait-19K: A Physically Grounded Synthetic Video Dataset for Gait Parameter Estimation}

\author{
    Soroush Mehraban$^{1,2,3}$ \quad
    Xin Lei Lin$^{1,2,3}$ \quad
    Vida Adeli$^{1,2,3}$\\
    Majid Mirmehdi$^{4}$ \quad
    Amirhossein Dadashzadeh$^{4}$ \quad
    Clint Hansen$^{5}$\\
    Andrea Iaboni$^{1,2}$ \quad
    Babak Taati$^{1,2,3}$\\[4pt]
    $^{1}$KITE Research Institute\\
    $^{2}$University of Toronto \quad
    $^{3}$Vector Institute\\
    $^{4}$University of Bristol \quad
    $^{5}$Kiel University
}

\begin{document}
\maketitle
\begin{abstract}
Accurate estimation of clinically meaningful gait parameters from monocular video is important for scalable mobility assessment, yet progress is limited by the small scale, restricted viewpoints, and limited visual diversity of existing datasets. We introduce \textbf{\dataset{}}, a physically grounded synthetic video dataset containing 19,272 walking videos derived from 6,427 MoCap sequences across 437 subjects, with paired SMPL motion and annotations for six gait parameters. To construct the dataset, we develop \textbf{\pipeline{}}, which unifies heterogeneous MoCap recordings through SMPL and synthesizes diverse RGB walking videos under controllable viewpoints and scene appearances. We assess the generated videos for consistency with their conditioning gait kinematics and validate extracted gait events against force-platform measurements. Using \dataset{}, we benchmark direct RGB, pose-based, biomechanical, and human-mesh-recovery approaches and analyze viewpoint, training-data scale, and synthetic-to-real domain shift. We also introduce \textbf{\model{}} as a direct RGB reference model for estimating gait parameters. Synthetic supervision transfers effectively to real videos across both \model{} and a pose-based architecture, demonstrating utility across different representations. We further find that spatial gait parameters are more sensitive to visual domain shift and that improved HMR reconstruction alone does not necessarily translate to improved downstream gait estimation.
\end{abstract}    
\section{Introduction}

Accurate estimation of human gait parameters has become an increasingly important goal in computer vision, driven by its broad impact in healthcare and mobility assessment. Gait features such as walking speed, cadence, step length, step width, stooped posture, and arm swing serve as critical biomarkers in diagnosing neurological conditions like Parkinson’s disease~\cite{sabo2022estimating, jacobs2005can, morris1996stride}, predicting fall risk in older adults~\cite{adeli2023ambient, rodriguez2019spatial, callisaya2012risk}, and tracking rehabilitation progress~\cite{barthuly2012gait}. Traditionally, gait evaluation relies on inertial measurement units (IMUs)~\cite{tao2012gait, seel2014imu} or marker-based motion capture (MoCap) systems~\cite{kadaba1990measurement, davis1991gait}. Although these approaches offer high accuracy, they require laboratory environments, specialized equipment, and extensive subject preparation, which restricts their use in routine clinical workflows. Furthermore, gait patterns measured in controlled settings often diverge from those observed in daily life~\cite{renggli2020wearable}, highlighting the need for ambient and unobtrusive capture methods for robust assessment in natural, everyday settings.

\begin{figure*}[t]
  \centering
  \includegraphics[width=0.95\textwidth]{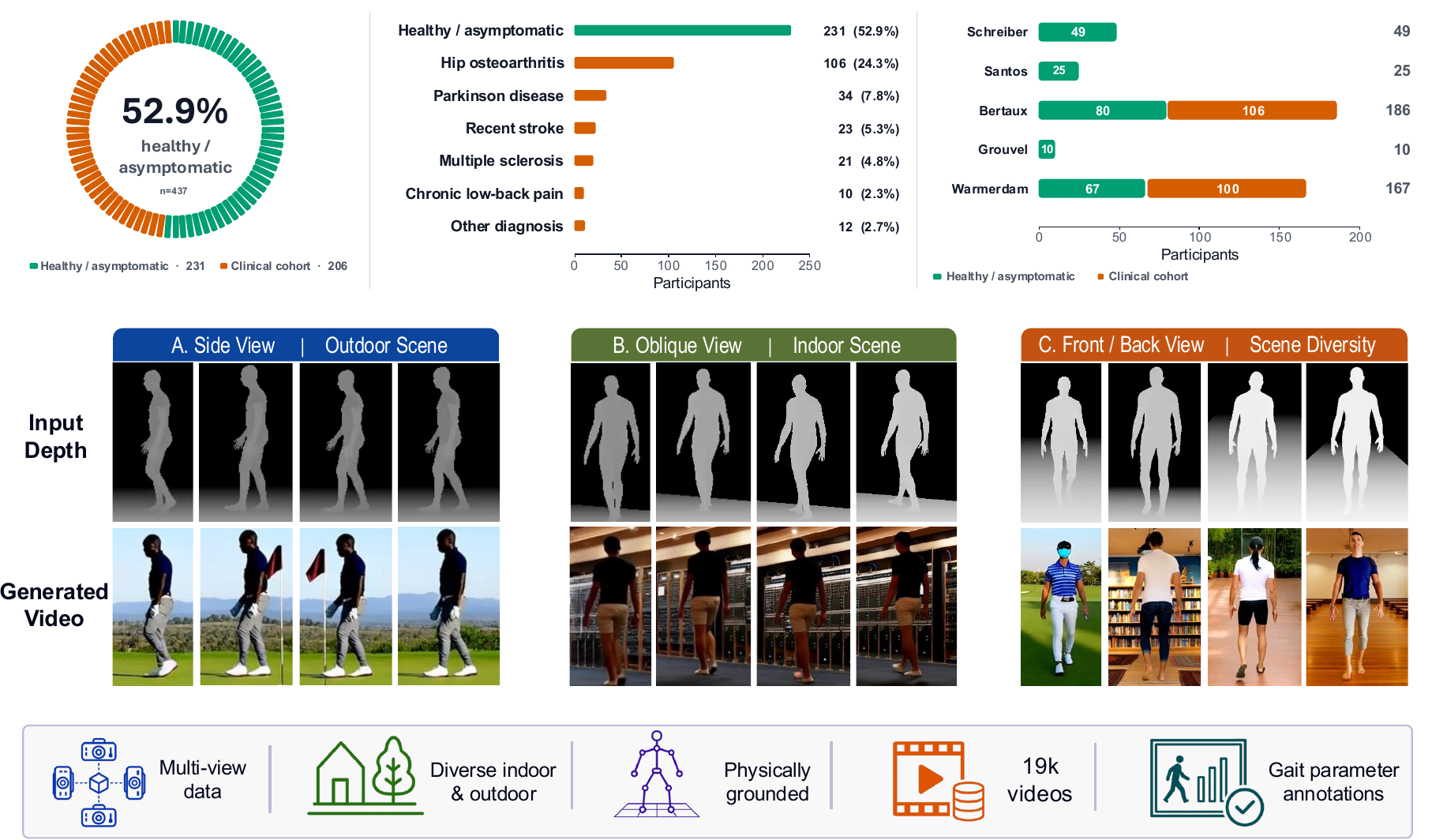}
\caption{
    \textbf{\dataset{} overview.}
    Source-cohort composition and representative depth-conditioned generations
    across viewpoints and diverse indoor/outdoor scenes.
    The dataset contains 19{,}273 videos from 437 subjects with paired 3D motion
    and six gait-parameter annotations.
    }
  \label{fig:teaser}
\end{figure*}

Vision-based gait assessment offers a promising path toward scalable and non-intrusive analysis, but progress is limited by the availability and scope of existing datasets. Estimating gait parameters from video typically requires supervision from force plates or MoCap systems, making large-scale collection expensive and difficult to share due to the identifiable biometric information contained in gait recordings. Consequently, existing datasets are often collected in a single controlled environment, with restricted camera viewpoints and limited variation in subject appearance. These constraints make it difficult to assess whether a method generalizes beyond the capture setup on which it was developed. They also make it challenging to isolate the effect of individual factors such as camera viewpoint, visual appearance, or scene variation, since these factors typically change together across datasets. As a result, current evaluations provide only limited insight into how different gait-estimation approaches behave under controlled distribution shifts and which aspects of the visual domain are most responsible for performance degradation.

To address these limitations, we introduce \textbf{\dataset{}}, a large-scale, physically grounded synthetic video dataset for gait parameter estimation, with motion derived from real MoCap recordings. \dataset{} contains
19,273 RGB walking videos derived from 6,427 MoCap sequences across 437
subjects, with paired SMPL motion and annotations for six clinically
relevant gait parameters. As summarized in~\cref{fig:teaser}, the
source data span five independent cohorts, including 231 healthy or
asymptomatic participants and 206 participants from clinical populations,
providing diversity in both gait characteristics and capture protocols.
Each source motion is observed under multiple camera configurations and
diverse visual appearances, allowing viewpoint and scene variation to be
studied independently of the underlying fitted motion.

To construct \dataset{}, we develop \textbf{\pipeline{}}, a synthetic
video-generation pipeline that converts heterogeneous MoCap recordings
into a common SMPL representation, renders them from controllable virtual
cameras, and uses depth-conditioned video diffusion to generate diverse
RGB walking videos across indoor and outdoor scenes. We validate the
generated videos for consistency with their conditioning gait kinematics
and independently validate the extracted gait events against force-platform
measurements. The resulting dataset enables controlled study of viewpoint,
training-data scale, and synthetic-to-real visual shift, while providing
a common benchmark for methods with substantially different intermediate
representations.

We evaluate human mesh recovery (HMR), biomechanical, pose-based, and direct RGB
approaches on a common real-world gait benchmark. As a direct RGB reference
model, we introduce \textbf{\model{}}, a Video-ViT-based model that predicts
gait parameters directly from monocular video. Training both \model{} and a
pose-based STT model~\cite{le2024learning} on \dataset{} shows that synthetic supervision transfers
effectively to real video across different representations. Our experiments
further characterize viewpoint dependence, scaling behavior, and
synthetic-to-real domain shift, and show that improved intermediate HMR
reconstruction alone does not necessarily translate to improved downstream
gait estimation.

\paragraph{Contributions.}
(1) We introduce \textbf{\dataset{}}, a physically grounded synthetic
gait-video dataset containing 19,272 videos from 6,427 walking sequences
across 437 subjects, with paired SMPL motion and six gait-parameter
annotations.
(2) We develop \textbf{\pipeline{}} to construct the dataset and validate
the generated motion consistency and gait-event annotations through
kinematic-fidelity and force-platform analyses.
(3) We establish a benchmark spanning direct RGB, pose-based,
biomechanical, and HMR approaches, and use \dataset{} to study viewpoint,
data scale, and synthetic-to-real transfer, with \textbf{\model{}} serving
as a direct RGB reference model.
\begin{figure*}[!t]
  \centering
  \includegraphics[width=\linewidth]{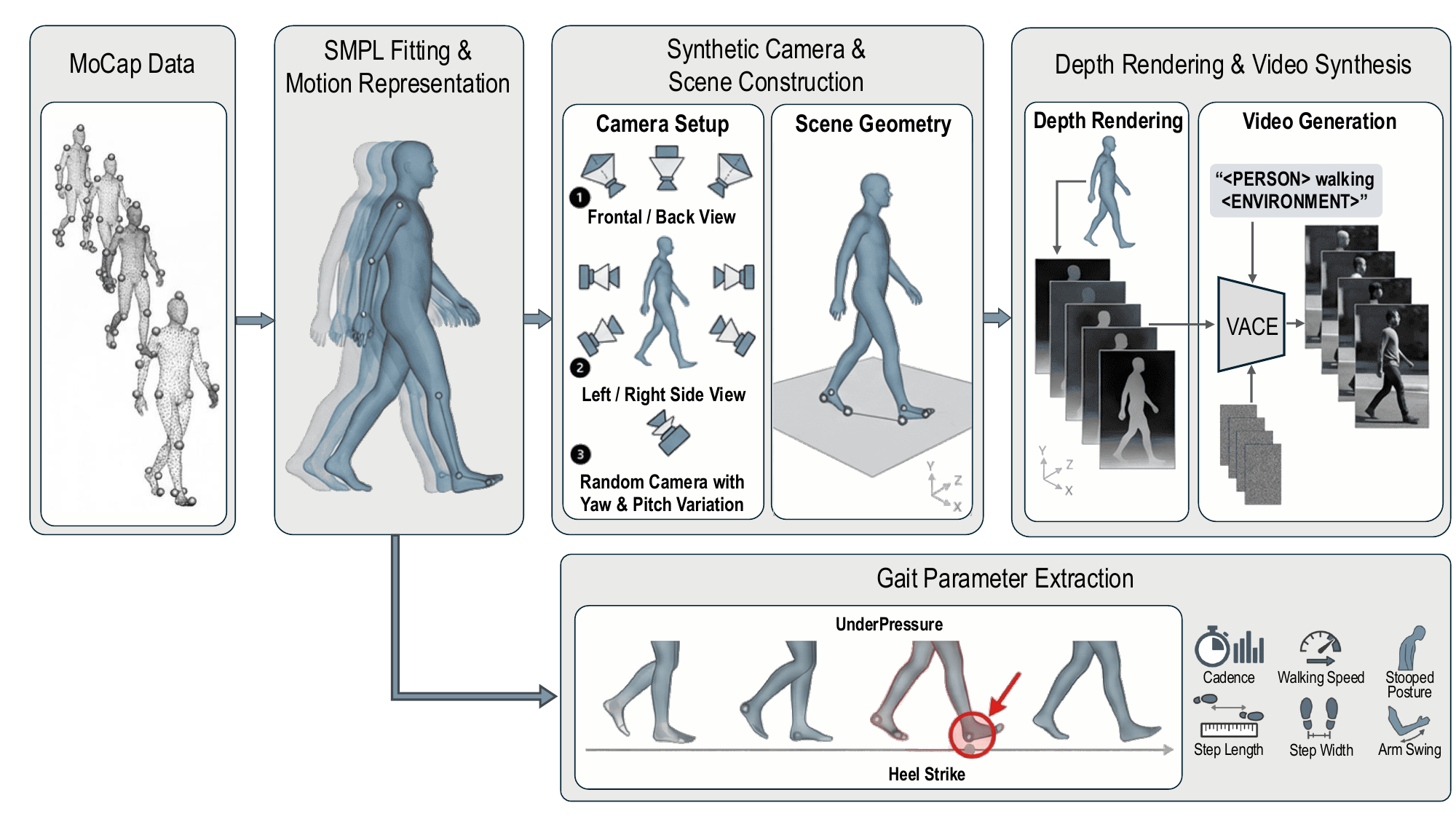}
  \caption{
    Overview of the synthetic video generation pipeline. 3D walking motions from MoCap data are fitted with SMPL, rendered into depth maps using controllable synthetic cameras and scene geometry, and used to condition a video diffusion model to generate diverse walking videos.
  }
  \label{fig:video-gen-pipeline}
\end{figure*}

\section{Related Work}

\subsection{Video-based Gait Parameter Estimation}

Vision-based gait analysis methods broadly fall into general-purpose human-motion reconstruction pipelines and task-specific gait-estimation models.

\noindent\textbf{Human mesh recovery--based methods.}
HMR methods estimate 3D body pose and shape from images~\cite{goel2023humans, patel2025camerahmr} or videos~\cite{wang2025prompthmr, mehraban2025fasthmr, shin2024wham}. Gait events and spatiotemporal parameters can then be derived from the reconstructed motion, for example using UnderPressure~\cite{mourot2022underpressure}. However, these models are primarily optimized for human reconstruction rather than downstream gait accuracy.

\noindent\textbf{Task-specific gait estimation.}
Other approaches directly target gait quantities from visual representations. Pose2Gait~\cite{malin2023pose2gait}, Kidziński \emph{et al.}~\cite{kidzinski2020deep}, and STT~\cite{le2024learning} estimate gait parameters from 2D pose trajectories, while Transforming Gait~\cite{cotton2022transforming} operates on estimated 3D joint trajectories. Biomechanical pipelines such as PBL~\cite{peiffer2025portable} and OpenCap Monocular~\cite{gilon2026opencap} further combine monocular reconstruction with biomechanical modeling. These methods span substantially different intermediate representations and objectives, motivating comparison based on the resulting gait measurements rather than reconstruction quality alone.

\subsection{Gait Datasets and Evaluation}

Existing gait datasets with synchronized video and motion or biomechanical measurements remain relatively small and capture-specific. GPJATK~\cite{kwolek2019calibrated} provides synchronized MoCap and calibrated multi-view RGB, while prior task-specific studies rely on dedicated dementia~\cite{malin2023pose2gait}, cerebral-palsy~\cite{kidzinski2020deep}, or instrumented gait-laboratory cohorts~\cite{cotton2022transforming}. Such datasets provide valuable real-world measurements but make it difficult to vary viewpoint, appearance, or scene independently while preserving the underlying motion. \dataset{} complements them with large-scale RGB walking videos whose camera and visual conditions can be controlled independently of the fitted gait motion.

\subsection{Synthetic Human Motion and Video Data}

Synthetic data has enabled scalable supervision for human-motion understanding. SURREAL~\cite{varol2017learning} generates SMPL-based synthetic humans from MoCap motion, while AGORA~\cite{patel2021agora} and BEDLAM~\cite{black2023bedlam} increase diversity in appearance, clothing, scenes, and camera configurations. These datasets primarily target general human reconstruction rather than quantitative gait analysis.

Most closely related, Yamada \emph{et al.}~\cite{yamada2025utility} study synthetic musculoskeletal gait data for healthcare applications using projected pose representations from simulated gait. In contrast, \pipeline{} starts from real-world recorded MoCap across multiple cohorts and generates diverse RGB videos while retaining correspondence with fitted 3D motion and gait annotations, enabling controlled analysis of viewpoint and synthetic-to-real visual variation.
\section{\dataset{} Dataset}
\label{sec:dataset}

\begin{figure}[t]
    \centering
    \includegraphics[width=\columnwidth]{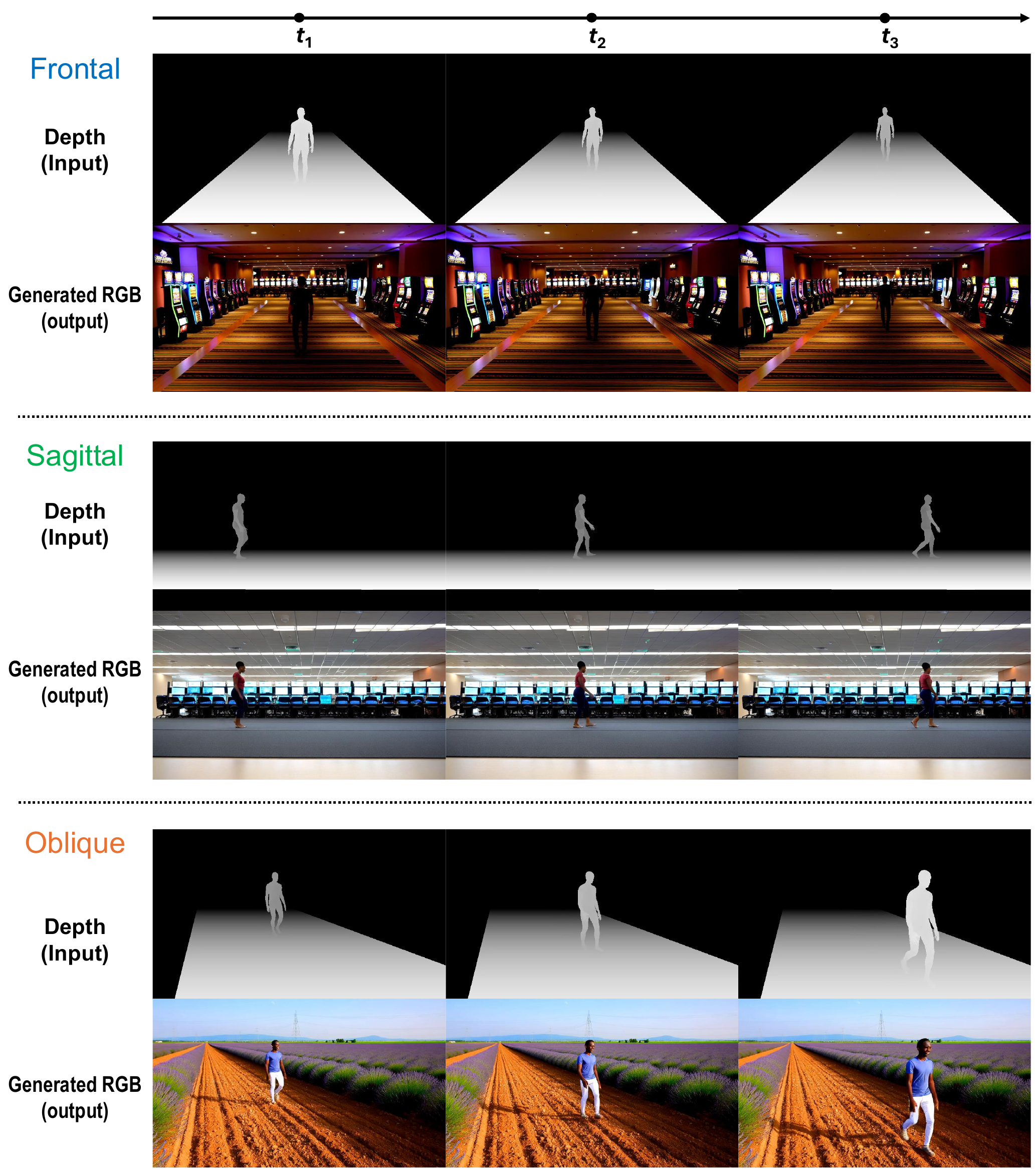}
    \caption{
        Qualitative samples from \dataset{} showing the diversity of generated
        subjects, environments, viewpoints, and walking motions.
        Each generated RGB video is paired with the corresponding SMPL motion
        and gait annotations.
    }
    \label{fig:synthgait-qualitative}
\end{figure}

\noindent\textbf{Dataset composition.}
\dataset{} contains 19,272 RGB walking videos derived from 6,427 unique
MoCap sequences comprising 671 minutes of walking from 437 subjects across
five public datasets
~\cite{schreiber2019multimodal,santos2022multi,bertaux2022gait,
grouvel2023dataset,warmerdam2022full}. As summarized in
\cref{fig:teaser}, the source cohorts include both healthy or
asymptomatic participants and multiple clinical populations. Each video
is paired with its fitted SMPL motion and annotations for six gait
parameters: cadence, walking speed, step length, step width, stooped
posture, and arm swing. \Cref{tab:dataset-stats} summarizes the
source-specific subject, sequence, and video counts.

\begin{table*}[t]
    \centering
    \caption{
    \textbf{Composition of \dataset{} and real-video evaluation data.}
    \dataset{} combines five public MoCap datasets; GPJATK is the
    independent real-video benchmark. Counts report subjects, sequences,
    and videos by viewpoint.
    }
    \label{tab:dataset-stats}

    \footnotesize
    \renewcommand{\arraystretch}{0.95}
    \setlength{\tabcolsep}{5pt}

    \begin{tabular*}{\textwidth}{
        @{\extracolsep{\fill}}
        lcccccc
        @{}
    }
    \toprule
    \multirow{2}{*}{\textbf{Dataset / motion source}}
    & \multirow{2}{*}{\textbf{\# Subjects}}
    & \multirow{2}{*}{\textbf{\# Sequences}}
    & \multicolumn{4}{c}{\textbf{\# Videos}} \\
    \cmidrule(lr){4-7}
    &
    &
    &
    \textbf{Front/Back}
    & \textbf{Sagittal}
    & \textbf{Oblique}
    & \textbf{Total} \\
    \midrule

    \multicolumn{7}{@{}l}{\textbf{\dataset{} composition}} \\
    Schreiber \& Moissenet~\cite{schreiber2019multimodal}
    & 49  & 917 & 917 & 918 & 918 & 2{,}752 \\
    Santos \emph{et al.}~\cite{santos2022multi}
    & 25  & 488 & 488 & 488 & 488 & 1{,}464 \\
    Bertaux \emph{et al.}~\cite{bertaux2022gait}
    & 186 & 4{,}442 & 4{,}441 & 4{,}442 & 4{,}436 & 13{,}319 \\
    Grouvel \emph{et al.}~\cite{grouvel2023dataset}
    & 10  & 82 & 82 & 82 & 82 & 246 \\
    Warmerdam \emph{et al.}~\cite{warmerdam2022full}
    & 167 & 497 & 497 & 497 & 497 & 1{,}491 \\
    \cmidrule(lr){1-7}
    \textbf{\dataset{} total}
    & \textbf{437} & \textbf{6{,}427}
    & \textbf{6{,}425} & \textbf{6{,}427}
    & \textbf{6{,}421} & \textbf{19{,}272} \\

    \midrule
    \multicolumn{7}{@{}l}{\textbf{Independent real-video evaluation}} \\
    GPJATK~\cite{kwolek2019calibrated}
    & 32 & 152 & 152 & 152 & 304 & 608 \\
    \bottomrule
    \end{tabular*}
\end{table*}

\noindent\textbf{Viewpoint and visual diversity.}
Each source motion is observed under front, back, sagittal, and oblique
camera configurations, with additional variation in camera pitch, subject
appearance, scene, and lighting. Because these factors can vary while the
fitted motion and gait labels remain fixed, \dataset{} enables controlled
analysis of viewpoint and visual-domain variation.
\Cref{fig:synthgait-qualitative} shows representative samples from the
same underlying walking sequence across multiple views and visual
conditions. We use an 80/20 subject-level training/validation split so
that all videos from a given subject remain within the same partition.
\section{\pipeline{}: Dataset Construction Pipeline}
\label{sec:gait2vid}

We propose \pipeline{}, a synthetic data-generation pipeline for constructing
\dataset{} from heterogeneous MoCap recordings. As illustrated in
\cref{fig:video-gen-pipeline}, \pipeline{} converts source motions into
a common SMPL representation, renders them from controllable virtual
cameras, synthesizes diverse RGB videos using depth-conditioned video
diffusion, and derives gait annotations from the fitted motion.

\noindent\textbf{Unified motion representation.}
The five source datasets use different marker layouts and joint
conventions, preventing their raw trajectories from being combined
directly. We therefore convert each dataset into a common
SMPL~\cite{loper2015smpl} representation, providing a consistent
full-body representation for video generation and gait annotation.

\noindent\textbf{Synthetic camera and scene construction.}
Given a fitted SMPL walking sequence, we render depth videos from front,
back, sagittal, and randomly sampled oblique cameras with varying pitch.
A planar ground surface is rendered beneath the walking trajectory to
provide a stable geometric reference during synthesis.

\noindent\textbf{Depth-conditioned video synthesis.}
The rendered depth sequence conditions Wan2.1-14B-VACE
~\cite{jiang2025vace}, together with a text prompt describing the subject
and environment. Prompts are sampled from 200 curated indoor and outdoor
scene templates, producing variation in subject appearance, background,
lighting, and recording context while conditioning on the fitted motion.

\noindent\textbf{Gait parameter extraction.}
Heel strikes are detected from the fitted SMPL motion using
UnderPressure~\cite{mourot2022underpressure}. Cadence is computed from
their timing; walking speed from pelvis displacement; step length and
width from forward and mediolateral foot displacement between successive
steps; stooped posture from normalized neck--pelvis forward displacement;
and arm swing from normalized wrist range along the walking direction.
Each generated video remains paired with gait labels derived from its
conditioning motion. Dataset-specific SMPL fitting, video synthesis, and
gait-parameter definitions are detailed in the supplementary material.
\section{Benchmark Protocol}

\subsection{Evaluation Dataset and Metrics}

\noindent\textbf{Real-world evaluation.}
We evaluate all methods on GPJATK~\cite{kwolek2019calibrated}, a synchronized RGB--MoCap gait dataset containing 152 walking sequences from 32 subjects. The dataset provides 608 RGB videos across four camera configurations: 152 side, 76 front, 76 back, and 304 oblique views. The RGB videos are used as model input, while the synchronized MoCap recordings are used to derive reference gait parameters using the same gait-annotation definitions as \dataset{}.

\noindent\textbf{Preferred-view protocol.}
Because different gait parameters are most observable from different camera directions, we define a fixed preferred-view protocol used for the main benchmark. Walking speed, step length, stooped posture, and arm swing are evaluated from the side view; step width is evaluated using front and back views; and cadence is evaluated across all views. We additionally report view-wise results to characterize sensitivity to camera viewpoint.

\noindent\textbf{Metrics.}
We report Pearson correlation ($r$) to measure how well each method preserves inter-sequence variation in each gait parameter. The overall correlation is obtained by averaging the six parameter-wise correlations using Fisher $z$-transformation. We additionally report mean absolute error (MAE) in the native units of each gait parameter in the supplementary material to assess absolute prediction accuracy.

\subsection{\model{}}
\begin{figure}[!t]
    \centering
    \includegraphics[width=\columnwidth]{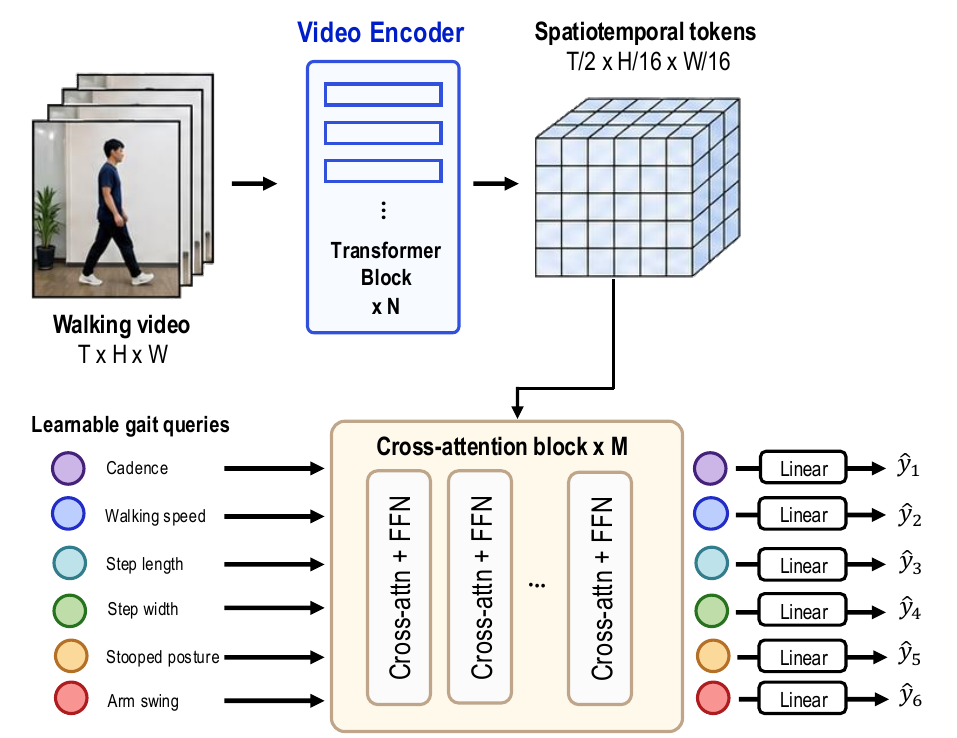}
    \caption{
        \textbf{\model{} overview.}
        An input walking video is encoded by a Video ViT into spatiotemporal tokens. A set of six learnable gait-query tokens cross-attends, and parameter-specific linear heads produce the final gait estimates.
    }
    \label{fig:\model{}-pipeline}
\end{figure}

We introduce \model{}, a direct RGB model for estimating gait parameters without an intermediate pose or mesh representation. As illustrated in~\cref{fig:\model{}-pipeline}, a V-JEPA2-initialized Video ViT encodes the walking video into spatiotemporal tokens. Six learnable gait queries, one for each gait parameter, independently cross-attend to the video representation through a lightweight decoder, and parameter-specific linear heads predict cadence, walking speed, step length, step width, stooped posture, and arm swing. The model is trained end-to-end using mean-squared error over normalized gait parameters. Architecture and training details are provided in the supplementary material.

\subsection{Comparison Methods}

We benchmark methods spanning different representations and training objectives. 
For general-purpose human mesh recovery, we evaluate WHAM~\cite{shin2024wham}, CameraHMR~\cite{patel2025camerahmr}, PromptHMR~\cite{wang2025prompthmr}, and FastHMR~\cite{mehraban2025fasthmr}. For each method, the predicted 3D body motion is converted to the six gait parameters using the same gait-feature extraction procedure, allowing downstream gait accuracy to be compared under a common protocol.

To provide a task-specific comparison with matched synthetic supervision, we additionally train STT~\cite{le2024learning} on \dataset{}. Unlike the HMR baselines, STT operates on pose trajectories and is explicitly optimized for gait-parameter estimation, providing a complementary control for separating the effect of task-specific supervision from the choice of intermediate representation.

We further consider the biomechanical pipelines PBL~\cite{peiffer2025portable} and OpenCap Monocular~\cite{gilon2026opencap}, which are evaluated using their released formulations. Additional adaptation details and representation-specific constraints are provided in the supplementary material.
\section{Results and Analysis}

\subsection{Synthetic Video Kinematic Fidelity}
\begin{table}[t]
\centering
\caption{
\textbf{Kinematic fidelity of VACE-generated GPJATK videos.}
Sapiens2 pose estimates are compared with projected fitted-SMPL motion for paired real and generated videos. Deltas are Generated $-$ Real; 95\% CIs are obtained by sequence-level bootstrap.
}
\label{tab:vace-kinematic-fidelity}
\resizebox{\columnwidth}{!}{
\begin{tabular}{lcccc}
\toprule
Metric & Real & Generated & $\Delta$ & 95\% CI \\
\midrule
BBox-NMPJPE (\%) $\downarrow$
    & 3.41 & 2.32 & -1.09 & [-1.19, -0.98] \\
Lower-body NMPJPE (\%) $\downarrow$
    & 2.90 & 2.20 & -0.70 & [-0.79, -0.61] \\
Lower-body velocity error $\downarrow$
    & 0.74 & 0.76 & +0.02 & [-0.01, +0.04] \\
Knee-angle MAE ($^\circ$) $\downarrow$
    & 8.71 & 4.86 & -3.86 & [-4.11, -3.62] \\
\bottomrule
\end{tabular}}
\end{table}

We assess whether RGB synthesis preserves the conditioning motion by projecting each fitted SMPL sequence into the calibrated camera and comparing it with Sapiens2~\cite{khirodkar2026sapiens2} 2D pose estimates from the corresponding RGB video. Confidence intervals are obtained by bootstrapping walking sequences, keeping synchronized views grouped.

As shown in~\cref{tab:vace-kinematic-fidelity}, generated videos have lower pose and knee-angle errors than the corresponding real videos, likely due to cleaner visual conditions and reduced clothing-induced ambiguity, while velocity error is nearly unchanged. We therefore find no evidence that synthesis degrades adherence to the conditioning motion. Since fitted SMPL is not independent marker-level ground truth, this analysis measures motion consistency rather than absolute biomechanical accuracy.

\subsection{Gait Annotation Validation}
\begin{table}[t]
\centering
\caption{
\textbf{UnderPressure heel-strike validation against force-platform measurements.}
Timing errors are reported in 30-FPS SMPL frames over force-platform-observed contacts.
}
\label{tab:heelstrike-validation}
\begin{tabular}{cccc}
\toprule
Events & MAE (frames) $\downarrow$ & $\leq$3 frames $\uparrow$ & $\leq$5 frames $\uparrow$ \\
\midrule
6,092 & 2.31 & 82.3\% & 91.7\% \\
\bottomrule
\end{tabular}
\end{table}

\pipeline{} gait annotations rely on heel strikes detected from fitted SMPL motion using UnderPressure~\cite{mourot2022underpressure}. We validate their timing against force-platform contacts from the Bertaux \etal~\cite{bertaux2022gait} subset. Because the force plates cover only part of the walkway, they provide an independent reference for observed contacts, while UnderPressure remains necessary for the full sequence.

Across 6,092 force-platform-observed heel strikes, UnderPressure achieves a mean absolute timing error of 2.31 30-FPS SMPL frames ($\approx77$\,ms), with 82.3\% and 91.7\% localized within 3 and 5 frames, respectively (\cref{tab:heelstrike-validation}).

\subsection{Real-World Gait Estimation Benchmark}
\begin{table}[t]
    \caption{
    Preferred-view comparison across gait parameters. We report Pearson correlation ($r$).
    \textbf{Cad}: cadence, \textbf{W.Speed}: walking speed,
    \textbf{Step Len}: step length, \textbf{Step Wid}: step width,
    \textbf{Stoop Post}: stooped posture, \textbf{Arm Swing}: arm swing.
    \textbf{Avg} denotes the Fisher $z$-transformed average across parameters.
$\dagger$ denotes STT trained on SynthGait-19K; -- indicates unsupported outputs.
    }
    \label{tab:main-comparison}
    \centering
    \scriptsize
    \renewcommand{\arraystretch}{1.05}
    \begin{tabular*}{\columnwidth}{@{\extracolsep{\fill}}lccccccc@{}}
    \toprule
    \textbf{Method}
    & \makecell{\textbf{Cad}}
    & \makecell{\textbf{W.}\\\textbf{Speed}}
    & \makecell{\textbf{Step}\\\textbf{Len}}
    & \makecell{\textbf{Step}\\\textbf{Wid}}
    & \makecell{\textbf{Stoop}\\\textbf{Post}}
    & \makecell{\textbf{Arm}\\\textbf{Swing}}
    & \makecell{\textbf{Avg}} \\
    \midrule

    \multicolumn{8}{@{}l}{\textbf{HMR}} \\
    WHAM~\cite{shin2024wham}      & 0.85 & 0.82 & 0.40 & \textbf{0.69} & 0.64 & 0.67 & 0.70 \\
    CameraHMR~\cite{patel2025camerahmr} & 0.71 & 0.65 & 0.10 & 0.35 & 0.39 & 0.91 & 0.59 \\
    PromptHMR~\cite{wang2025prompthmr} & 0.87 & 0.61 & 0.13 & 0.48 & 0.63 & 0.90 & 0.67 \\
    FastHMR~\cite{mehraban2025fasthmr}   & 0.80 & 0.63 & -0.07 & 0.18 & 0.02 & \textbf{0.93} & 0.54 \\

    \midrule
    \multicolumn{8}{@{}l}{\textbf{Biomechanical}} \\
    PBL~\cite{peiffer2025portable}
        & 0.47 & 0.77 & 0.38 & 0.50 & 0.73 & 0.63 & 0.60 \\
    OpenCap-M~\cite{gilon2026opencap}
        & 0.76 & 0.87 & 0.62 & 0.47 & 0.58 & 0.65 & 0.68 \\

    \midrule
    \multicolumn{8}{@{}l}{\textbf{2D Pose-based}} \\
    STT~\cite{le2024learning}
        & 0.20 & 0.67 & -- & -- & -- & -- & -- \\
    STT$^\dagger$
        & 0.91 & 0.85 & \textbf{0.73} & 0.67 & 0.70 & 0.92 & 0.82 \\

    \midrule
    \multicolumn{8}{@{}l}{\textbf{Video-based}} \\
    \makecell[l]{\textbf{GaitXFormer}}
        & \textbf{0.94} & \textbf{0.88} & 0.67 & 0.67
        & \textbf{0.75} & 0.91 & \textbf{0.84} \\

    \bottomrule
    \end{tabular*}
\end{table}
\Cref{tab:main-comparison} compares gait-estimation approaches spanning HMR, biomechanical, 2D pose-based, and direct video-based formulations. The HMR and biomechanical approaches first recover an intermediate body representation from which gait parameters are derived, whereas STT and \model{} are optimized specifically for gait estimation. Across the off-the-shelf HMR and biomechanical approaches, performance varies substantially by gait parameter: OpenCap Monocular performs strongly for walking speed, WHAM for step width, and FastHMR for arm swing.

To evaluate whether the benefit of \dataset{} extends beyond
\model{}, we additionally train the STT architecture on
\dataset{} using the same six gait-parameter annotations. The released STT model was trained on side-view videos from a cerebral-palsy cohort and therefore exhibits limited cross-dataset transfer to GPJATK, particularly for cadence ($r=0.20$), while its walking-speed correlation is $0.67$. After training on \dataset{}, STT$^\dagger$ reaches a Fisher-averaged correlation of 0.82 on real GPJATK, compared with 0.84 for \model{}, and achieves the strongest step-length result among the evaluated methods. This substantial improvement indicates that \dataset{} provides useful supervision for a substantially different pose-based architecture, rather than benefiting only \model{}. At the same time, the parameter-wise differences across methods highlight that no single representation is uniformly optimal for all gait quantities.

\begin{figure}[!t]
    \centering
    \includegraphics[width=\columnwidth]{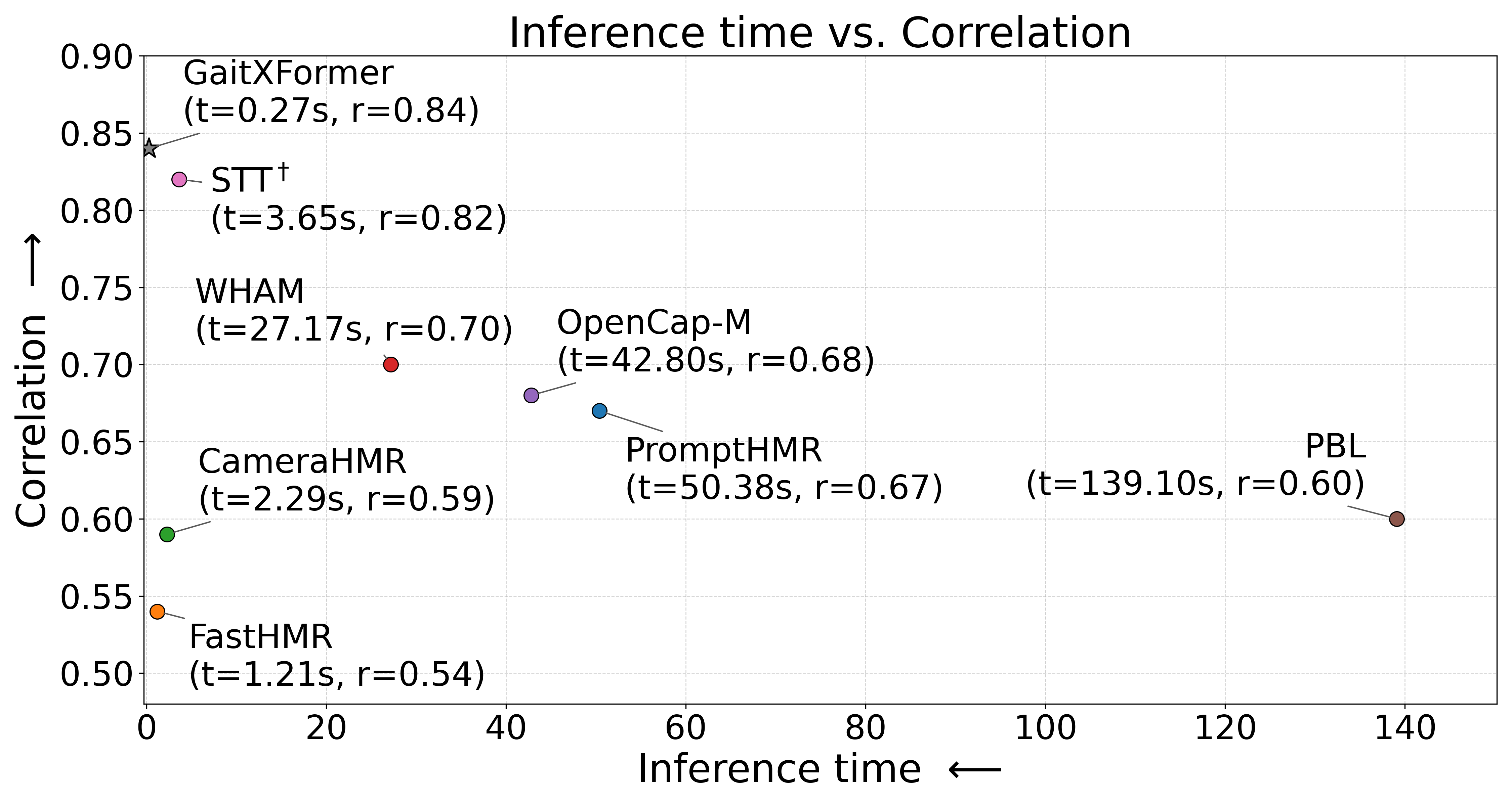}
    \caption{
\textbf{Accuracy--runtime trade-off.}
Pearson correlation versus inference time for a 5-second walking clip on a single NVIDIA RTX3090 GPU.
}
\label{fig:runtime-comparison}
\end{figure}

\noindent\textbf{Inference efficiency.}
\model{} processes a 5-second walking clip in 0.27\,s on a single RTX3090, compared with 1.21--139.10\,s for the evaluated baselines, including required preprocessing. As shown in~\cref{fig:runtime-comparison}, it achieves the most favorable accuracy--runtime trade-off. STT itself requires only 0.32\,s, but additionally relies on OpenPose preprocessing.

\subsection{Synthetic-to-Real Transfer Analysis}
\begin{table}[t]
\caption{
\textbf{Synthetic-to-real transfer analysis.}
GPJATK-VACE contains VACE-generated videos using GPJATK motions and camera viewpoints. \textbf{Syn.} indicates synthetic RGB input, and \textbf{Avg} denotes the Fisher $z$-transformed average across gait parameters.
}
\label{tab:synthetic-real-transfer}
\centering
\scriptsize
\renewcommand{\arraystretch}{1.05}
\resizebox{\columnwidth}{!}{
\begin{tabular}{@{}lcccccccc@{}}
\toprule
\textbf{Eval.}
& \makecell{\textbf{Syn.}}
& \makecell{\textbf{Cad}}
& \makecell{\textbf{W.}\\\textbf{Spd}}
& \makecell{\textbf{Step}\\\textbf{Len}}
& \makecell{\textbf{Step}\\\textbf{Wid}}
& \makecell{\textbf{Stoop}\\\textbf{Post}}
& \makecell{\textbf{Arm}\\\textbf{Swing}}
& \makecell{\textbf{Avg}} \\
\midrule

\multicolumn{9}{@{}l}{\textit{GaitXFormer}} \\
Synth. val
& \checkmark
& 0.71 & 0.96 & 0.93 & 0.92 & 0.96 & 0.93 & 0.92 \\
GPJATK-VACE
& \checkmark
& 0.93 & 0.89 & 0.74 & 0.77 & 0.80 & 0.93 & 0.87 \\
GPJATK
& $\times$
& 0.94 & 0.88 & 0.67 & 0.67 & 0.75 & 0.91 & 0.84 \\

\addlinespace[2pt]
\midrule

\multicolumn{9}{@{}l}{\textit{WHAM}} \\
GPJATK-VACE
& \checkmark
& 0.82 & 0.79 & 0.40 & 0.56 & 0.79 & 0.61 & 0.68 \\
GPJATK
& $\times$
& 0.85 & 0.82 & 0.40 & 0.69 & 0.64 & 0.67 & 0.70 \\

\bottomrule
\end{tabular}
}
\end{table}

\pipeline{} allows us to examine the effect of visual-domain change while retaining the same source gait motions and camera configurations. We therefore compare performance on GPJATK-VACE, generated from GPJATK motion, with performance on the corresponding real GPJATK domain. We additionally report performance on the held-out \dataset{} validation set to distinguish transfer to unseen synthetic motion from transfer to real imagery.

\model{} exhibits a modest decrease in Fisher-averaged correlation from 0.87 on GPJATK-VACE to 0.84 on real GPJATK. Cadence, walking speed, and arm swing remain nearly unchanged, whereas larger reductions occur for step length and step width, suggesting that spatial gait quantities are more sensitive to the synthetic-to-real appearance shift.

WHAM shows a different pattern. Its overall correlation remains similar across generated and real GPJATK (0.68 vs.\ 0.70), but the effect varies considerably across parameters. Step width improves from 0.56 to 0.69 on real video, while stooped-posture correlation decreases from 0.79 to 0.64. These results indicate that synthetic-to-real sensitivity is not a uniform property of the dataset, but depends on both the estimated gait quantity and the intermediate representation used by the model.

\subsection{Effect of \dataset{} Supervision on HMR}
\begin{table}[t]
\caption{
\textbf{Effect of SynthGait supervision on WHAM.}
Reconstruction and gait correlations are evaluated on real GPJATK. Adapt. denotes WHAM with the \dataset{}-trained output adapter.
}
\label{tab:wham-adaptation}
\centering
\scriptsize
\renewcommand{\arraystretch}{1.05}
\begin{tabular*}{\columnwidth}{@{\extracolsep{\fill}}lccc@{}}
\toprule
\textbf{Metric} & \textbf{WHAM} & \textbf{Adapt.} & \textbf{Change} \\
\midrule

\multicolumn{4}{@{}l}{\textbf{SMPL reconstruction} $\downarrow$} \\
Body-pose error ($^\circ$)
    & 10.54 & 9.76 & $-7.43\%$ \\
Pelvis MPJPE (mm)
    & 46.93 & 45.63 & $-2.78\%$ \\
PA-MPJPE (mm)
    & 33.28 & 32.44 & $-2.53\%$ \\

\midrule
\multicolumn{4}{@{}l}{\textbf{Gait correlation} $\uparrow$} \\
Walking speed
    & 0.826 & 0.842 & +0.017 \\
Step length
    & 0.402 & 0.442 & +0.040 \\
Step width
    & 0.687 & 0.666 & -0.021 \\
Arm swing
    & 0.669 & 0.651 & -0.017 \\
Avg
    & 0.705 & 0.708 & +0.004 \\
\bottomrule
\end{tabular*}
\end{table}

To examine whether \dataset{} supervision can improve an HMR-based gait pipeline, we adapt WHAM using the paired SMPL motion available in \dataset{}. We keep the released WHAM RGB-to-motion network fixed and train a lightweight temporal output adapter that refines its predicted pose, shape, and trajectory scale. We then evaluate both the resulting motion reconstruction and the gait parameters derived from the adapted predictions.

As shown in~\cref{tab:wham-adaptation}, \dataset{} supervision improves WHAM reconstruction on real GPJATK, reducing body-pose error by 7.43\%, pelvis MPJPE by 2.78\%, and PA-MPJPE by 2.53\%. These reconstruction gains, however, do not translate uniformly to gait estimation. The Fisher-averaged correlation changes only modestly from 0.7045 to 0.7080 ($\Delta r=0.0035$), with a participant-bootstrap 95\% confidence interval of $[-0.0068,\,0.0166]$. Walking speed and step length improve, while step width and arm swing decrease. This result highlights that improved human-motion reconstruction does not necessarily imply improved accuracy for derived gait quantities.

\subsection{Controlled Analyses Enabled by \dataset{}}

\noindent\textbf{Sensitivity to camera viewpoint.}
\dataset{} provides controlled training data across diverse camera configurations while preserving the underlying gait motion. To assess whether the resulting model remains robust to viewpoint changes in real video, we evaluate \model{} on the multi-view GPJATK benchmark. As shown in~\cref{tab:viewpoint-corr}, cadence is highly stable across camera configurations, whereas other gait parameters exhibit stronger viewpoint dependence. Stooped posture and arm swing are best estimated from the side view, while step width benefits substantially from front and back views. Walking speed and step length remain relatively robust across views, with the highest correlations observed for the oblique cameras. These results characterize the remaining viewpoint sensitivity after multi-view synthetic training and motivate the parameter-specific preferred-view protocol used in our main benchmark.

\begin{table}[t]
\caption{Viewpoint-wise Pearson correlation ($r$, $\uparrow$) of \model{} on GPJATK. Numbers in parentheses denote the number of test videos.}
\label{tab:viewpoint-corr}
\centering
\footnotesize
\renewcommand{\arraystretch}{1.05}
\begin{tabular*}{\columnwidth}{@{\extracolsep{\fill}}lcccc@{}}
\toprule
\textbf{Gait Feature} &
\makecell{\textbf{Side}\\\textbf{(152)}} &
\makecell{\textbf{Front}\\\textbf{(76)}} &
\makecell{\textbf{Back}\\\textbf{(76)}} &
\makecell{\textbf{Oblique}\\\textbf{(304)}} \\
\midrule
Cadence    & \textbf{0.95} & 0.94 & 0.94 & 0.92 \\
Walking Speed   & 0.88 & 0.87 & 0.87 & \textbf{0.91} \\
Step Length   & 0.67 & 0.62 & 0.64 & \textbf{0.70} \\
Step Width   & 0.50 & 0.63 & \textbf{0.70} & 0.48 \\
Stooped Posture & \textbf{0.75} & 0.26 & 0.24 & 0.59 \\
Arm Swing  & \textbf{0.91} & 0.77 & 0.78 & \textbf{0.91} \\
\bottomrule
\end{tabular*}
\end{table}

\noindent\textbf{Effect of synthetic training-data scale.}
\dataset{} also allows us to directly measure how increasing the amount of synthetic supervision affects transfer to real video. We train the same \model{} configuration using progressively larger subsets of \dataset{} and evaluate each model on real GPJATK. As shown in~\cref{tab:dataset-scale-ablation}, the Fisher-averaged correlation increases from 0.79 using 25\% of the training data to 0.84 using the full dataset. Most gait parameters improve as additional synthetic data are introduced, with particularly clear gains for cadence, step length, and step width. The continued improvement up to the full training set indicates that the scale of \dataset{} contributes directly to downstream real-world performance.

\begin{table}[t]
    \caption{Effect of training-data scale on gait-estimation performance in terms of Pearson correlation ($r$). \textbf{Avg} denotes the Fisher $z$-transformed average across gait parameters.}
    \label{tab:dataset-scale-ablation}
    \centering
    \scriptsize
    \renewcommand{\arraystretch}{1.05}
    \begin{tabular*}{\columnwidth}{@{\extracolsep{\fill}}lccccccc@{}}
    \toprule
    \textbf{Train}
    & \makecell{\textbf{Cad}}
    & \makecell{\textbf{W.}\\\textbf{Speed}}
    & \makecell{\textbf{Step}\\\textbf{Len}}
    & \makecell{\textbf{Step}\\\textbf{Wid}}
    & \makecell{\textbf{Stoop}\\\textbf{Post}}
    & \makecell{\textbf{Arm}\\\textbf{Swing}}
    & \makecell{\textbf{Avg}} \\
    \midrule
    25\%
    & 0.87 & 0.84 & 0.60 & 0.59 & \textbf{0.77} & 0.89 & 0.79 \\
    50\%
    & 0.92 & 0.84 & 0.57 & 0.62 & 0.72 & 0.91 & 0.80 \\
    75\%
    & 0.92 & 0.87 & 0.66 & 0.65 & 0.74 & \textbf{0.92} & 0.83 \\
    100\%
    & \textbf{0.94} & \textbf{0.88} & \textbf{0.67} & \textbf{0.67} & 0.75 & 0.91 & \textbf{0.84} \\
    \bottomrule
    \end{tabular*}
\end{table}

\subsection{External Clinical Transfer on PD4T}
\begin{table}[t]
\caption{
\textbf{UPDRS classification on PD4T} using frozen video encoders under leave-one-subject-out evaluation.
}
\label{tab:pd4t-transfer}
\centering
\footnotesize
\setlength{\tabcolsep}{8pt}
\renewcommand{\arraystretch}{1.05}
\begin{tabular}{@{}lccc@{}}
\toprule
\textbf{Encoder}
& \textbf{Acc.}
& \makecell{\textbf{Macro}\\\textbf{Recall}}
& \makecell{\textbf{Macro}\\\textbf{F1}} \\
\midrule
V-JEPA2
& 71.3 & 68.3 & 69.1 \\
GaitXFormer
& \textbf{76.0} & \textbf{71.3} & \textbf{73.3} \\
\bottomrule
\end{tabular}
\end{table}

We further evaluate transfer to PD4T~\cite{dadashzadeh2024pecop}, an independent dataset of Parkinson's-disease walking videos with UPDRS severity labels. First, we freeze the \model{} video encoder and train a lightweight classifier for UPDRS prediction, using the same protocol with the pretrained V-JEPA2 encoder as a baseline. Under leave-one-subject-out evaluation, the \dataset{}-trained \model{} representation improves accuracy from 71.3\% to 76.0\% and macro F1 from 69.1\% to 73.3\% (~\cref{tab:pd4t-transfer}), indicating that gait supervision produces features that transfer to clinical severity estimation.

\begin{table}[t]
\caption{
Patient-aware clinical-anchor analysis on PD4T. Predictions are averaged within each participant and UPDRS-score group before correlation analysis. Dir. denotes within-patient agreement with the predefined expected clinical direction, and SRM denotes standardized response mean.
}
\label{tab:pd4t-patient-aware}
\centering
\scriptsize
\renewcommand{\arraystretch}{1.05}
\resizebox{\columnwidth}{!}{
\begin{tabular}{@{}lcccc@{}}
\toprule
\textbf{Parameter}
& \makecell{\textbf{Spearman}\\$\boldsymbol{\rho}$}
& \makecell{\textbf{95\% CI}}
& \makecell{\textbf{Dir.}}
& \makecell{\textbf{SRM}} \\
\midrule
Walking Speed
& -0.83 & $[-0.88,-0.76]$ & 97.9\% & 1.37 \\
Step Length
& -0.83 & $[-0.88,-0.78]$ & 97.9\% & 1.39 \\
Arm Swing
& -0.75 & $[-0.84,-0.61]$ & 91.5\% & 1.24 \\
Stooped Posture
& 0.30 & $[0.14,0.47]$ & 80.9\% & 0.53 \\
\bottomrule
\end{tabular}
}
\end{table}
We additionally examine whether \model{}'s predicted gait quantities themselves vary consistently with clinical severity. To account for repeated observations, predictions are averaged within each participant and UPDRS-score group, with confidence intervals obtained by bootstrapping participants. As shown in~\cref{tab:pd4t-patient-aware}, walking speed, step length, and arm swing show strong negative associations with increasing UPDRS severity ($\rho=-0.83$, $-0.83$, and $-0.75$) and change in the expected direction in more than 91\% of within-participant comparisons. Stooped posture shows a weaker positive association ($\rho=0.30$), with 80.9\% directional agreement. Together, these results provide complementary evidence that representations and gait quantities learned from synthetic supervision transfer to unseen clinical videos.
\section{Conclusion}
We introduced \dataset{}, a large-scale synthetic video dataset for gait analysis constructed from heterogeneous MoCap recordings and paired with unified SMPL motion and six gait-parameter annotations. Beyond providing training data, \dataset{} enables controlled evaluation of viewpoint, visual-domain shift, and training-data scale. Our experiments show that synthetic supervision transfers effectively to real video for several gait parameters, while spatial quantities remain more sensitive to domain shift. Improved HMR reconstruction does not necessarily improve downstream gait
estimation, while predicted gait quantities show consistent associations
with clinical severity on PD4T.

Limitations include incomplete coverage of real-world variation,
particularly severe occlusion, assistive devices, and broader clothing
and clinical diversity. The fixed 5-s window also excludes longer-horizon
phenomena such as freezing of gait. Evaluation centers on GPJATK, with PD4T validation; broader cohorts
would better characterize generalization.
{
    \small
    \bibliographystyle{ieeenat_fullname}
    \bibliography{main}
}

\appendix
\clearpage

\begin{center}
    {\LARGE \textbf{Supplementary Material}}\\[6pt]
    {\large \dataset{}: A Physically Grounded Synthetic Video Dataset for Gait Parameter Estimation}
\end{center}

\vspace{1em}

\section{Physically-Grounded Video Generation}
\label{appendix:video-gen}
\subsection{Motion Sources and SMPL Conversion}

We obtain walking motion from motion capture (MoCap) datasets. MoCap datasets employ heterogeneous joint placement conventions depending on the capture system. 
To ensure consistency, each dataset is independently interpolated and converted into a unified joint representation aligned with the SMPL kinematic structure.


\subsection{Camera Configuration and Viewpoint Sampling}
For each SMPL walking sequence, we render depth videos using three camera configurations. 
The first configuration places the camera in a frontal or back view relative to the walking direction, where only the camera pitch is varied. 
The second configuration places the camera in a left or right side view, again varying only the pitch angle. 
The third configuration samples a random viewpoint by uniformly rotating the camera yaw in the range $[0^\circ, 360^\circ]$.

For all three configurations, the camera pitch is randomly sampled from the range $[-5^\circ, 45^\circ]$, enabling viewpoints ranging from near-horizontal to moderately elevated perspectives.
\subsection{Depth Rendering with Planar Ground}

\begin{figure}[t]
    \centering
    \includegraphics[width=\columnwidth]{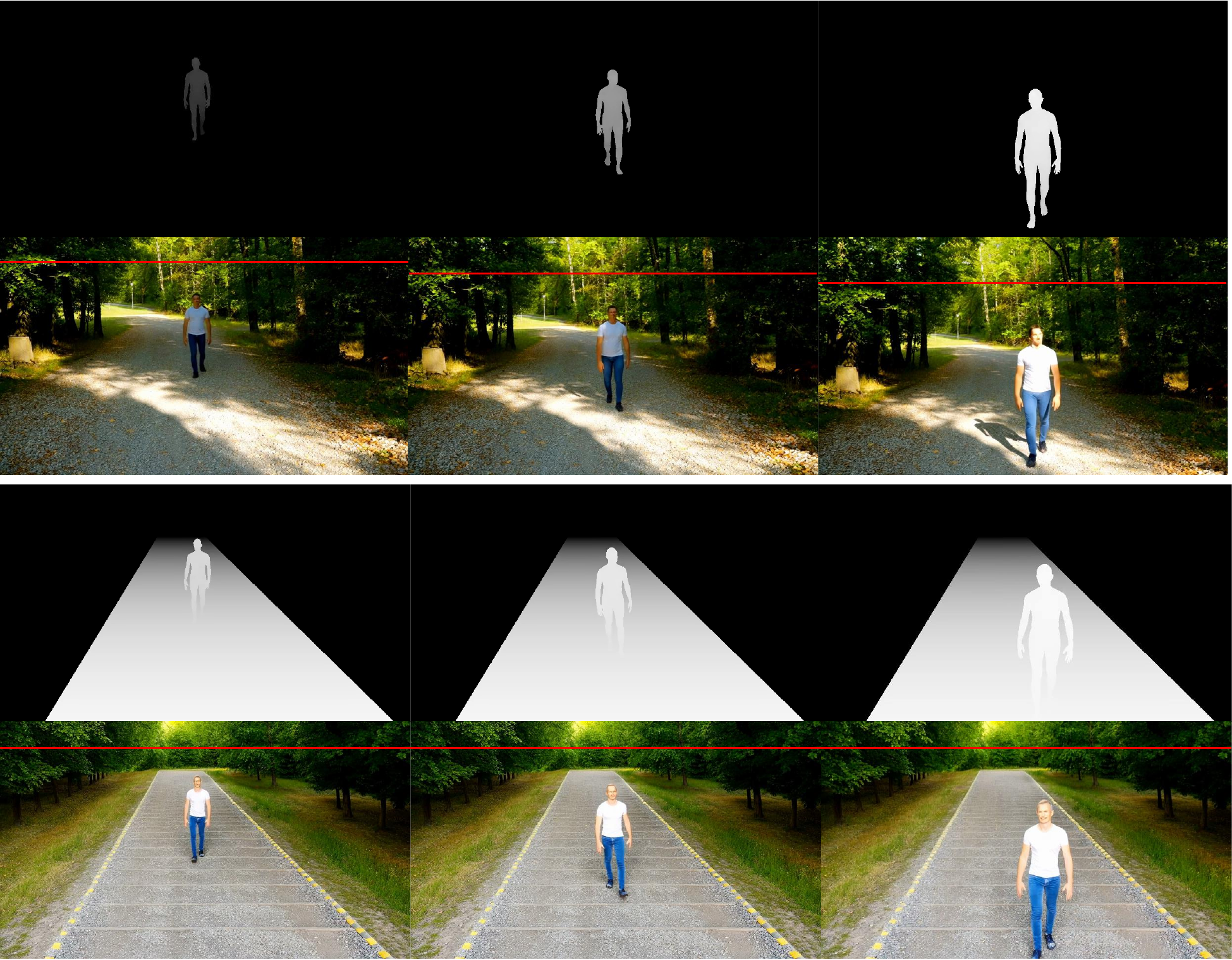}
    \caption{
        Effect of adding a synthetic ground plane on camera stability, where the red line marks the horizon and shows how missing geometry induces spurious camera pitch and scale changes across frames.
    }
    \label{fig:effect-of-plane}
\end{figure}

Rendering depth maps of the human body alone provides limited environmental context and can lead to undesired zooming and frame-to-frame camera instability during video synthesis. 
As shown in~\cref{fig:effect-of-plane}, missing scene geometry can induce spurious changes in camera pitch and scale across frames. 
To provide a stable geometric reference, we place a synthetic planar ground beneath the walking trajectory during depth rendering. 
The plane extends along the walking direction with a randomly sampled width.

Since the depth of the planar ground remains static throughout the sequence, this setup implicitly enforces a static camera configuration during RGB video generation. 
This design significantly improves temporal consistency and prevents camera drift artifacts in the synthesized videos.

\subsection{RGB Video Synthesis with VACE}
Given a rendered depth video, we synthesize RGB walking videos using the Wan2.1-14B-VACE~\cite{jiang2025vace} model. 
Video generation is conditioned on both the depth sequence and a text prompt describing the subject and environment. We use TeaCache~\cite{liu2025timestep} to increase generation speed. 
The prompt specifies the subject's gender and nationality, along with an environment description sampled from a curated set of indoor and outdoor scenes.

\noindent\textbf{Generation compute.}
Video synthesis is an offline, one-time dataset construction cost.
The production campaign used Wan2.1-VACE-14B at $480\times832$
resolution with 81 frames, 50 denoising steps, and TeaCache
(threshold 0.3) on NVIDIA L40S GPUs. Generation was divided into
48 independent shards and executed in parallel on up to 48 GPUs,
requiring approximately 3,131 allocated L40S-hours.

The generation campaign produced a larger candidate pool prior to
dataset curation, from which the final 19,273 videos from 6,427
motions comprising \dataset{} were retained. Consequently, the
reported generation compute also includes samples that were not
retained in the final dataset. This cost reflects the particular
Wan2.1-VACE-14B configuration used for this release rather than an
intrinsic computational requirement of \pipeline{}; the video-generation
backbone is modular and can benefit from faster generation models
and inference techniques.

\subsection{Gait Parameter Extraction}
Gait-parameter annotations are computed directly from
the fitted SMPL walking sequences.
We apply the UnderPressure model~\cite{mourot2022underpressure} to detect heel-strike events from the motion sequence. 
Using the detected heel strikes together with 3D joint trajectories of the hips, feet, and neck, we compute a set of physically meaningful 3D gait features.

These features serve as supervision signals for the downstream gait estimation task.

\paragraph{Walking Speed.}
Walking speed measures the average forward velocity of the subject during locomotion.
Assuming that walking occurs predominantly along the forward axis, we compute walking speed using the displacement of the pelvis joint between the first and last detected heel strikes, normalized by elapsed time:
\begin{equation}
v = \frac{\left| z_{\mathrm{pelvis}}(t_{\mathrm{last}}) - z_{\mathrm{pelvis}}(t_{\mathrm{first}}) \right|}{\frac{t_{\mathrm{last}} - t_{\mathrm{first}}}{\mathrm{fps}}},
\end{equation}
where $z_{\mathrm{pelvis}}(t)$ denotes the pelvis position along the forward axis at frame $t$, and $\mathrm{fps}$ is the frame rate.

\paragraph{Cadence.}
Cadence quantifies the temporal rhythm of walking and is defined as the number of steps per minute.
Given a sequence of detected heel strikes, cadence is computed as:
\begin{equation}
\mathrm{Cadence} = \frac{(N_{\mathrm{HS}} - 1)}{\frac{t_{\mathrm{last}} - t_{\mathrm{first}}}{\mathrm{fps}}} \times 60,
\end{equation}
where $N_{\mathrm{HS}}$ is the total number of detected heel strikes.

\paragraph{Step Length.}
Step length represents the forward distance covered between consecutive steps.
For each heel strike, we record the forward-axis position of the stepping foot.
Step length is computed as:
\begin{equation}
\ell_i = \left| z_{i+1} - z_i \right|,
\end{equation}
where $z_i$ is the forward position of the foot at the $i$-th heel strike.
The reported step length is the mean over all steps:
\begin{equation}
\overline{\ell} = \frac{1}{N-1} \sum_{i=1}^{N-1} \ell_i.
\end{equation}

\paragraph{Step Width.}
Step width measures the mediolateral spacing between consecutive steps and reflects balance during walking.
At each heel strike, the lateral-axis position of the stepping foot is extracted.
Step width is computed as:
\begin{equation}
w_i = \left| x_{i+1} - x_i \right|,
\end{equation}
where $x_i$ denotes the mediolateral foot position at the $i$-th heel strike.
The final step width is given by:
\begin{equation}
\overline{w} = \frac{1}{N-1} \sum_{i=1}^{N-1} w_i.
\end{equation}

\paragraph{Stooped Posture.}
Stooped posture characterizes the forward-leaning configuration of the upper body during walking.
It is computed as the horizontal distance between the neck and pelvis joints, normalized by leg length:
\begin{equation}
s(t) = \frac{z_{\mathrm{neck}}(t) - z_{\mathrm{pelvis}}(t)}{L_{\mathrm{leg}}},
\end{equation}
where $L_{\mathrm{leg}}$ denotes the subject’s leg length.
The reported stooped posture is the temporal mean:
\begin{equation}
\overline{s} = \frac{1}{T} \sum_{t=1}^{T} s(t).
\end{equation}

\paragraph{Arm Swing.}
Arm swing captures the amplitude of arm motion along the direction of walking.
Joint positions are first centered at the pelvis to remove global translation.
For each wrist, the normalized range of forward motion is computed as:
\begin{equation}
\begin{aligned}
a_{\mathrm{R}} &=
\frac{\max_t z_{\mathrm{rwrist}}(t) - \min_t z_{\mathrm{rwrist}}(t)}
     {L_{\mathrm{leg}}}, \\
a_{\mathrm{L}} &=
\frac{\max_t z_{\mathrm{lwrist}}(t) - \min_t z_{\mathrm{lwrist}}(t)}
     {L_{\mathrm{leg}}}.
\end{aligned}
\end{equation}
The final arm swing measure is:
\begin{equation}
\overline{a} = \frac{1}{2}\left(a_{\mathrm{R}} + a_{\mathrm{L}}\right).
\end{equation}

\subsection{Force-Platform Validation of Heel-Strike Annotations}
\label{sec:force_validation}

We validate the heel-strike events used for gait-parameter extraction against force-platform measurements available in the Bertaux \emph{et al.}~\cite{bertaux2022gait} dataset. All retained force-platform contacts were recorded at an analog sampling rate of 1,000\,Hz. For each vertical-force channel, contact was considered active when $|F_z|\geq20$\,N. Within-plate gaps of at most 0.03\,s were bridged, after which contact segments shorter than 0.10\,s were discarded. The first above-threshold sample of each retained segment defined the force-platform contact onset. Contacts assigned to the same foot with onset times within 0.25\,s were merged, retaining the earliest onset.

All processed SMPL sequences are represented at 30\,FPS. Each force-platform onset was therefore mapped to the corresponding processed frame as
\begin{equation}
f_{\mathrm{GRF}}
=
\operatorname{round}
\left(
t_{\mathrm{onset}} r_{\mathrm{SMPL}}
\right)
-
f_{\mathrm{offset}}.
\end{equation}
where $t_{\mathrm{onset}}$ is the force-platform onset time, $r_{\mathrm{SMPL}}=30$\,FPS, and $f_{\mathrm{offset}}$ denotes the starting-frame offset of the processed sequence.

UnderPressure~\cite{mourot2022underpressure} heel strikes are defined by $0\!\rightarrow\!1$ transitions of its estimated heel-contact signal. To evaluate temporal localization independently of laterality, left- and right-foot events were pooled and each force-platform onset was matched to its nearest UnderPressure event. The signed timing error is defined as
\begin{equation}
e_f = f_{\mathrm{UP}} - f_{\mathrm{GRF}},
\end{equation}
such that negative values indicate that UnderPressure detects contact before the force signal crosses the 20\,N threshold.

Of 3,245 evaluated trials, 3,053 contain at least one force-platform contact within the processed SMPL interval, yielding 6,092 heel strikes. The resulting mean absolute timing error is 2.305 frames ($\approx76.8$\,ms), with a median absolute error of 2 frames and an RMSE of 3.045 frames. Overall, 82.3\% and 91.7\% of heel strikes are localized within 3 and 5 frames, respectively. Because the force platforms instrument only part of the walkway, this evaluation measures timing accuracy for plate-observed contacts rather than recall over the complete walking sequence.
\begin{figure*}[!t]
    \centering
    \includegraphics[width=\linewidth]{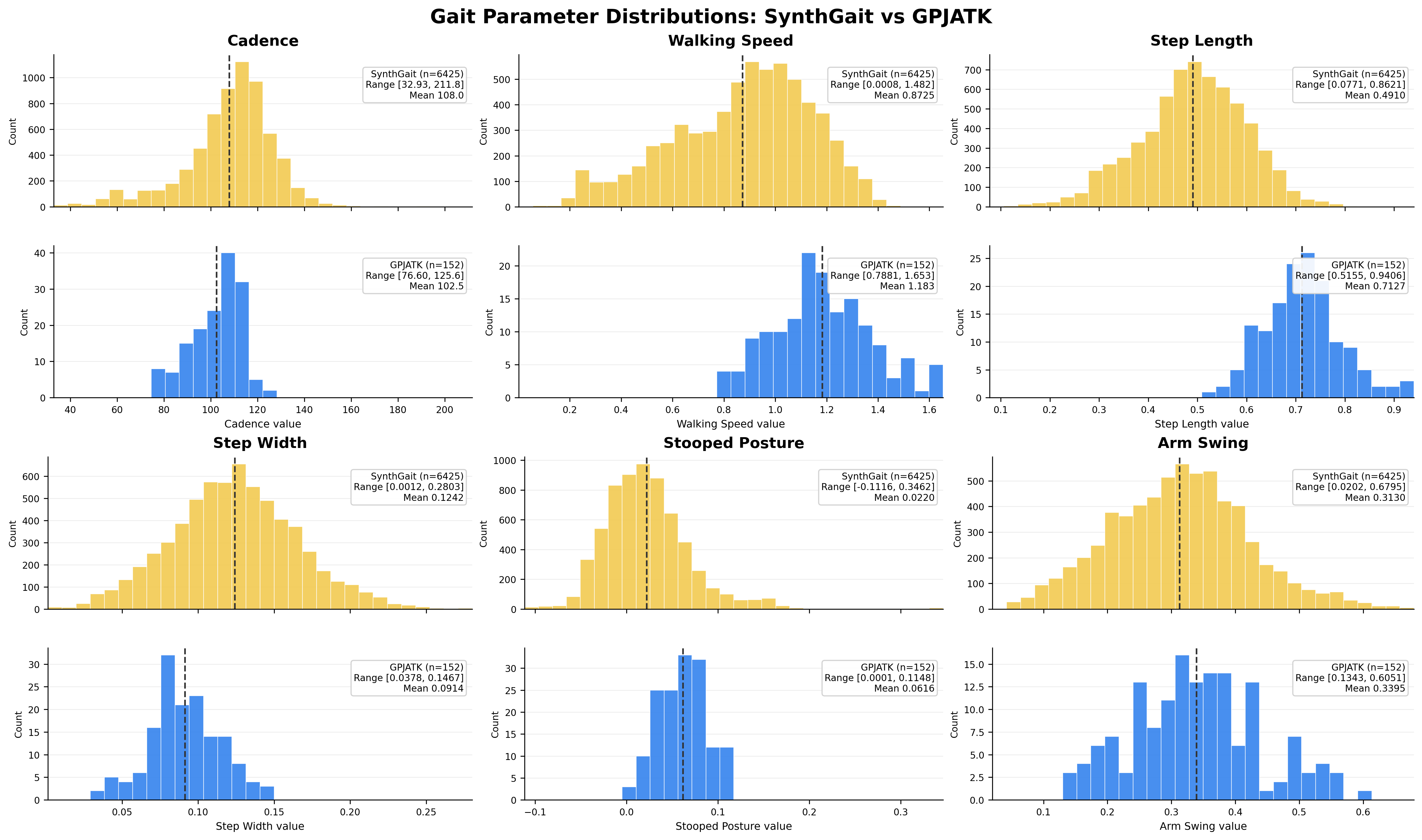}
    \caption{
    Distribution comparison of gait parameters in \dataset{} and GPJATK.
    For each gait parameter, the top histogram shows the proposed \dataset{} training distribution and the bottom histogram shows the GPJATK evaluation distribution. Dashed vertical lines indicate dataset means.
    }
    \label{fig:synthgait-gpjatk-distributions}
\end{figure*}

\section{Gait Parameter Distribution in \dataset{} and GPJATK}
Figure~\ref{fig:synthgait-gpjatk-distributions} compares the distributions of the six gait parameters in \dataset{} and GPJATK. Overall, \dataset{} provides substantially broader coverage across most gait parameters, while GPJATK forms a smaller and more concentrated evaluation distribution. This is expected since \dataset{} is designed to expose the model to diverse walking patterns during training, whereas GPJATK is a real-world evaluation dataset with a limited number of recorded trials.

Importantly, the GPJATK distributions largely fall within regions covered by \dataset{}, indicating that the synthetic training set captures gait variations relevant to the real evaluation data. At the same time, \dataset{} extends beyond GPJATK in several parameters, such as cadence, step width, stooped posture, and arm swing, which encourages the model to learn a broader representation rather than overfitting to the narrower range of the target evaluation set.
\section{Additional \model{} Details}
\label{sec:\model{}-details}

\subsection{Architecture and Training}

\noindent\textbf{Input processing.}
Input walking videos are represented by a fixed temporal window of
$T=32$ frames sampled from a 5-second clip. After person-centric
cropping, each frame is resized to $H=256$, $W=192$. Videos shorter
than the required temporal window are zero-padded. All gait parameters
are normalized using z-score normalization computed from the
\dataset{} training set.

\noindent\textbf{Video encoder.}
Given an input video
$X\in\mathbb{R}^{T\times H\times W}$, \model{} first encodes it
using a Video Vision Transformer:
\begin{equation}
    Z = \mathcal{E}(X) \in \mathbb{R}^{L\times D},
    \qquad
    L =
    \frac{T}{2}
    \frac{H}{16}
    \frac{W}{16},
\end{equation}
where $D$ is the embedding dimension and $L$ is the number of
spatiotemporal tokens. The encoder is initialized from V-JEPA2 and
contains 24 transformer layers with embedding dimension $D=1024$.
It uses a temporal stride of 2 and a spatial patch size of
$16\times16$. All encoder layers are fine-tuned during training on
\dataset{}.

\noindent\textbf{Gait-query decoder.}
To obtain parameter-specific representations, we introduce
$N=6$ learnable gait-query tokens,
\begin{equation}
    Q=\left\{q^{(i)}\right\}_{i=1}^{N},
    \qquad
    q^{(i)}\in\mathbb{R}^{D},
\end{equation}
with one query corresponding to each of cadence, walking speed,
step length, step width, stooped posture, and arm swing. The queries
attend to the video representation through a three-layer decoder:
\begin{equation}
    H = \mathrm{Decoder}(Q,Z)
    \in \mathbb{R}^{N\times D}.
\end{equation}
Each decoder layer consists of cross-attention from the gait queries
to the encoded video tokens followed by a feed-forward network. We
omit self-attention among the gait queries so that each query directly
retrieves evidence relevant to its corresponding gait parameter.

\noindent\textbf{Prediction heads.}
Each decoded representation $h^{(i)}$ is mapped independently to a
scalar gait estimate using a parameter-specific linear head:
\begin{equation}
    \hat{y}^{(i)}
    =
    w_i^\top h^{(i)} + b_i,
    \qquad i=1,\ldots,N.
\end{equation}
Stacking the individual predictions gives
$\hat{\mathbf{y}}\in\mathbb{R}^{N}$.

\noindent\textbf{Training objective.}
We train \model{} end-to-end using mean-squared error over the
normalized gait parameters:
\begin{equation}
    \mathcal{L}_{\mathrm{MSE}}
    =
    \frac{1}{BN}
    \sum_{b=1}^{B}
    \sum_{i=1}^{N}
    \left(
        \hat{y}_{b,i}-y_{b,i}
    \right)^2 ,
\end{equation}
where $B$ denotes the batch size. We use AdamW with a learning rate
of $1\times10^{-4}$ and train for 100 epochs with a global batch size
of 16 across four NVIDIA L40S GPUs.
\section{Benchmark Method Details}
\label{sec:benchmark_details}

\noindent\textbf{STT.}
We evaluate both the released STT~\cite{le2024learning} model and
an STT variant trained on \dataset{}. The released model was
trained on side-view videos from a cerebral-palsy cohort and
predicts cadence and walking speed. For STT$^\dagger$, we train
the STT architecture from scratch on \dataset{}, using six
independent regressors corresponding to the six gait parameters.
Each model operates on 81-frame, 16-FPS sequences of 25 BODY\_25
2D joints extracted using RTMPose. All model parameters are
trained using Adam with a learning rate of $6\times10^{-4}$,
a batch size of 128, and normalized mean-squared error, with
early stopping based on validation error.

\noindent\textbf{PBL.}
We evaluate PBL~\cite{peiffer2025portable} using its released formulation. We do not fine-tune its MeTRAbs pose-estimation component~\cite{sarandi2020metrabs} on \dataset{} because PBL relies on the dense 87-landmark BML-MoVi convention~\cite{ghorbani2021movi}, whereas \dataset{} provides SMPL-based motion annotations, and no validated correspondence between these landmark definitions was available.

\noindent\textbf{OpenCap Monocular.}
OpenCap Monocular~\cite{gilon2026opencap} builds on WHAM~\cite{shin2024wham} as its monocular 3D human-motion estimator and subsequently refines the predicted motion through camera and biomechanical optimization. We evaluate the released OpenCap Monocular pipeline without additional adaptation. In our separate \dataset{} adaptation experiment on WHAM, improved SMPL reconstruction produced only a negligible change in downstream gait correlation, suggesting that improving the underlying HMR representation alone does not necessarily translate to improved gait estimation.

\noindent\textbf{WHAM adaptation.}
To examine whether \dataset{} supervision can improve an HMR-based gait pipeline, we adapt the released WHAM~\cite{shin2024wham} model while keeping its RGB feature extractor, motion encoder/decoder, trajectory network, contact prediction, and root-orientation prediction frozen. We attach a lightweight temporal output adapter ($\sim$1.08M trainable parameters) to WHAM's final SMPL predictions. The adapter refines body pose and shape and applies a bounded scale correction to the predicted translation displacement. Training uses released-WHAM predictions from \dataset{} RGB videos paired with the corresponding fitted-SMPL motion. Because the constituent motion datasets are highly imbalanced, batches are source-balanced, and PCGrad~\cite{yu2020gradient} is used to reduce conflicting pose gradients across motion sources. The training objective supervises SMPL pose, joint and bone geometry, shape, and traveled distance. The adapted model is then evaluated both for SMPL reconstruction on GPJATK-VACE and for downstream gait estimation on real GPJATK using the same gait-extraction protocol as the released WHAM model.
\section{Details of the PD4T experiments}
\subsection{Parkinson's severity estimation}

PD4T contains 418 annotated walking trials from 30 participants. Since each trial may contain multiple walking rounds and turns, we extract straight-walking segments, resulting in 1,666 video clips. For this experiment, we remove the \model{} gait-query decoder, freeze the video encoder, and train a lightweight MLP classifier to predict UPDRS severity. As a baseline, we apply the same frozen-encoder and classifier protocol to the pretrained V-JEPA2 encoder. Both representations are evaluated using leave-one-subject-out cross-validation.

\begin{figure}[t]
    \centering
    \includegraphics[width=\linewidth]{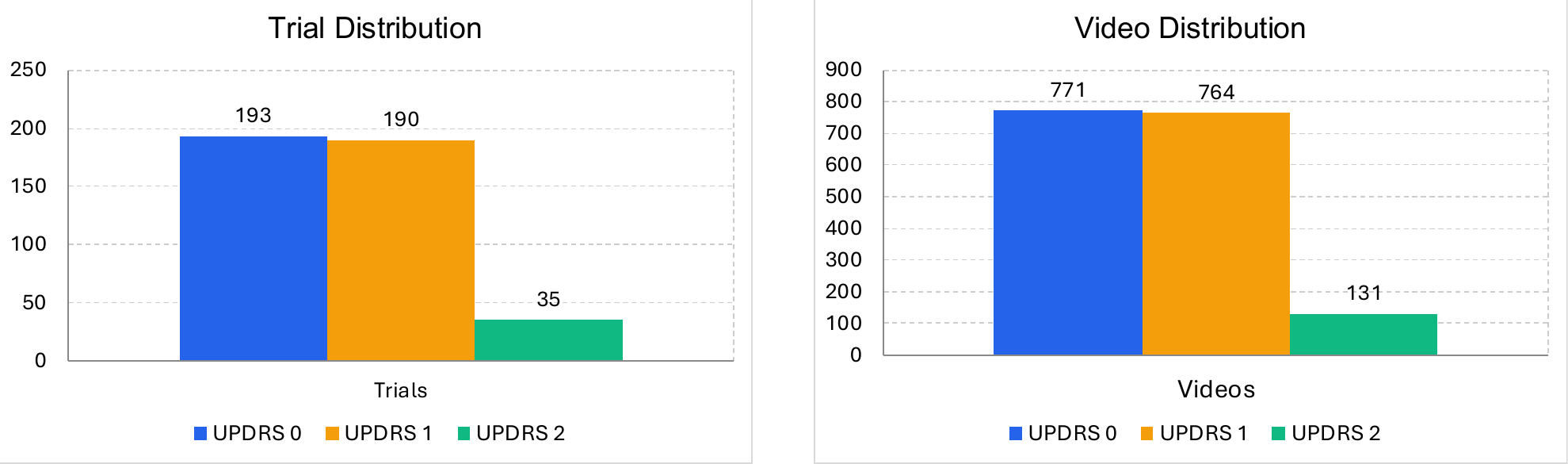}
    \caption{
    Distribution of PD4T samples across UPDRS severity classes. The left panel shows the number of trials per class, and the right panel shows the number of videos per class. The dataset is relatively balanced for UPDRS 0 and UPDRS 1, while UPDRS 2 is substantially underrepresented.
    }
    \label{fig:pd4t-class-distribution}
\end{figure}

\begin{figure}[t]
    \centering
    \includegraphics[width=\linewidth]{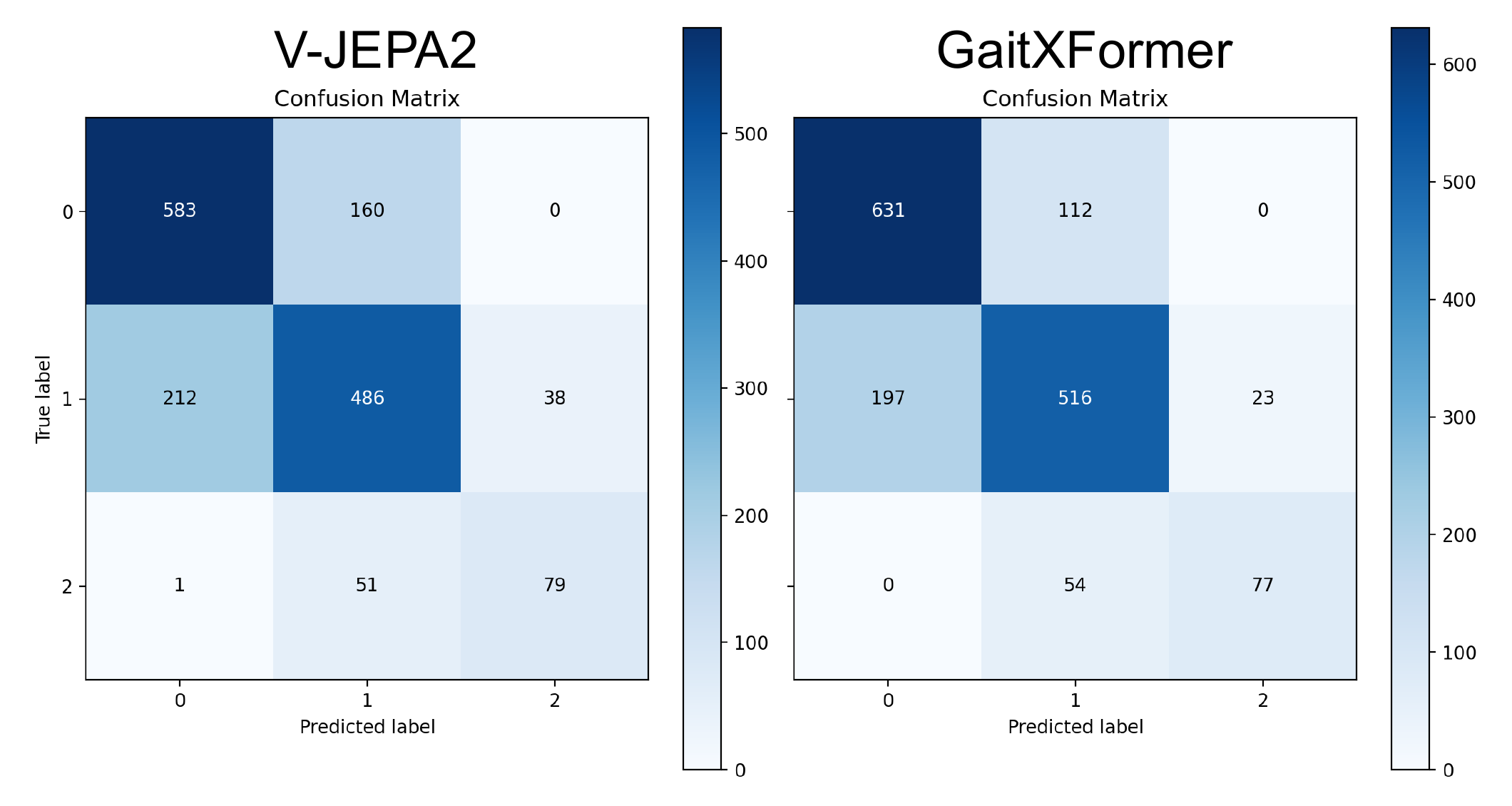}
    \caption{
    Confusion matrices for PD4T severity classification using frozen V-JEPA2 and \model{} backbones. Rows denote ground-truth labels and columns denote predicted labels. Compared with V-JEPA2, \model{} produces more diagonal predictions, particularly reducing confusion between classes 0 and 1, while most remaining errors occur between adjacent severity classes.
    }
    \label{fig:pd4t-confusion-matrices}
\end{figure}

Figures~\ref{fig:pd4t-class-distribution} and~\ref{fig:pd4t-confusion-matrices} provide additional details on this classification setting. Figure~\ref{fig:pd4t-class-distribution} shows that the dataset is fairly balanced between UPDRS 0 and UPDRS 1 at both the trial and video levels, whereas UPDRS 2 has substantially fewer samples. This class imbalance helps explain the trends observed in the confusion matrices in Figure~\ref{fig:pd4t-confusion-matrices}, where both models show stronger performance on the more frequent classes and comparatively weaker recognition of the underrepresented UPDRS 2 class. Despite this challenge, \model{} yields a cleaner diagonal structure and reduced confusion between adjacent classes compared with the frozen V-JEPA2 backbone.

\subsection{Patient-Aware Clinical-Anchor Evaluation}
For each participant, videos with the same clinical score were grouped together before analysis. We refer to each such group as a \emph{patient-score group}; for example, all videos from participant 001 with score 0 form one group, and all videos from participant 001 with score 1 form another group. Gait predictions were averaged within each group before computing correlations with clinical score. This avoids treating multiple videos from the same participant as independent samples.

Confidence intervals were estimated using patient-level cluster bootstrap resampling. In each bootstrap iteration, patients were sampled with replacement, all rows belonging to the sampled patients were retained, and the statistic of interest was recomputed. The reported 95\% confidence interval corresponds to the 2.5th and 97.5th percentiles across bootstrap iterations. This preserves the dependency structure among videos and score groups from the same participant.

For the within-patient directional analysis, adjacent clinical-score states from the same participant were compared. For a gait feature $f$, the change was computed as
\[
\Delta f = f_{\mathrm{higher\ score}} - f_{\mathrm{lower\ score}}.
\]
For features expected to decrease with worsening severity, such as walking speed, step length, and arm swing, the expected sign was set to $-1$. For stooped posture, which was expected to increase, the expected sign was set to $+1$. A pair was counted as changing in the expected direction when
\[
s_f \Delta f > 0,
\]
where $s_f$ is the expected sign for feature $f$. The effect size was computed as the standardized response mean,
\[
\mathrm{SRM}_f =
\frac{\mathrm{mean}(s_f \Delta f)}
{\mathrm{SD}(s_f \Delta f)}.
\]
\section{Additional Experimental Analysis}
\subsection{Mean Absolute Error Analysis}
\label{appendix:mae-analysis}

Table~\ref{tab:mae-appendix} reports preferred-view MAE in the
native units of each gait parameter. The expanded comparison
shows that correlation and absolute error capture complementary
aspects of performance. STT$^\dagger$ achieves particularly low
error for cadence and arm swing, while PBL and OpenCap-M obtain
the lowest walking-speed and step-length errors, respectively.
HMR-based methods remain competitive for stooped posture.
Together with the correlation results in the main paper, these
results highlight that no single representation is uniformly
best across all gait parameters.

\begin{table}[t]
    \caption{
    Preferred-view comparison using Mean Absolute Error (MAE).
    Lower is better. Cadence is reported in steps/min, walking
    speed in m/s, and step length and step width in meters;
    stooped posture and arm swing are dimensionless normalized
    quantities. $\dagger$ denotes STT trained on SynthGait-19K;
    -- indicates unsupported outputs.
    }
    \label{tab:mae-appendix}
    \centering
    \scriptsize
    \setlength{\tabcolsep}{2.0pt}
    \renewcommand{\arraystretch}{1.05}
    \resizebox{\columnwidth}{!}{
    \begin{tabular}{lcccccc}
    \toprule
    \textbf{Method}
    & \makecell{\textbf{Cad}}
    & \makecell{\textbf{W.}\\\textbf{Speed}}
    & \makecell{\textbf{Step}\\\textbf{Len}}
    & \makecell{\textbf{Step}\\\textbf{Wid}}
    & \makecell{\textbf{Stoop}\\\textbf{Post}}
    & \makecell{\textbf{Arm}\\\textbf{Swing}} \\
    \midrule

    \multicolumn{7}{l}{\textbf{HMR}} \\
    WHAM
    & 7.95 & 0.24 & 0.19 & 0.02 & 0.05 & 0.22 \\
    CameraHMR
    & 12.14 & 0.54 & 0.10 & 0.03 & 0.05 & 0.18 \\
    PromptHMR
    & 7.81 & 0.67 & 0.11 & 0.03 & \textbf{0.03} & 0.11 \\
    FastHMR
    & 10.41 & 0.59 & 0.12 & 0.03 & 0.05 & 0.13 \\

    \midrule
    \multicolumn{7}{l}{\textbf{Biomechanical}} \\
    PBL~\cite{peiffer2025portable}
    & 15.66 & \textbf{0.09} & 0.08 & 0.02 & 0.04 & 0.29 \\
    OpenCap-M~\cite{gilon2026opencap}
    & 7.56 & 0.10 & \textbf{0.07} & 0.04 & 0.12 & 0.12 \\

    \midrule
    \multicolumn{7}{l}{\textbf{2D Pose-based}} \\
    STT~\cite{le2024learning}
    & 13.99 & 0.11 & -- & -- & -- & -- \\
    STT$^\dagger$
    & \textbf{4.21} & 0.16 & 0.14 & \textbf{0.02} & 0.06 & \textbf{0.05} \\

    \midrule
    \multicolumn{7}{l}{\textbf{Video-based}} \\
    \textbf{GaitXFormer (ours)}
    & 9.97 & 0.22 & 0.19 & 0.02 & 0.08 & 0.07 \\

    \bottomrule
    \end{tabular}
    }
\end{table}

\subsection{Viewpoint Sensitivity Analysis}
\begin{table}[t]
    \caption{View-averaged comparison across gait parameters using Pearson correlation ($r$). 
    For each gait parameter, correlations are first computed separately for each camera view and then averaged across views using Fisher $z$-transformation. 
    \textbf{Avg} denotes the Fisher $z$-transformed average across gait parameters.}
    \label{tab:view-avg-comparison}
    \centering
    \scriptsize
    \renewcommand{\arraystretch}{1.05}
    \begin{tabular*}{\columnwidth}{@{\extracolsep{\fill}}lccccccc@{}}
    \toprule
    \textbf{Method}
    & \makecell{\textbf{Cad}}
    & \makecell{\textbf{W.}\\\textbf{Speed}}
    & \makecell{\textbf{Step}\\\textbf{Len}}
    & \makecell{\textbf{Step}\\\textbf{Wid}}
    & \makecell{\textbf{Stoop}\\\textbf{Post}}
    & \makecell{\textbf{Arm}\\\textbf{Swing}}
    & \makecell{\textbf{Avg}} \\
    \midrule
    \multicolumn{8}{@{}l}{\textbf{HMR}} \\
    WHAM & 0.85 & 0.77 & 0.34 & 0.55 & 0.46 & 0.72 & 0.65 \\
    CameraHMR & 0.71 & 0.44 & -0.16 & 0.35 & 0.42  & 0.88 & 0.51 \\
    PromptHMR & 0.87  & 0.37 & -0.20  & 0.46  & \textbf{0.64} & 0.89  & 0.60 \\
    FastHMR   & 0.80  & 0.44 & -0.20 & 0.33 & 0.12 & 0.91 & 0.51 \\
    \midrule
    \multicolumn{8}{@{}l}{\textbf{Video-based}} \\
    \makecell[l]{\textbf{GaitXFormer}}
              & \textbf{0.94}  & \textbf{0.88} & \textbf{0.66} & \textbf{0.58} & 0.49 & \textbf{0.86} & \textbf{0.79} \\
    \bottomrule
    \end{tabular*}
\end{table}
\begin{table}[t]
    \caption{Viewpoint-wise Pearson correlation ($r$, $\uparrow$) of WHAM on GPJATK. Numbers in parentheses denote the number of test videos.}
    \label{tab:wham-view-analysis}
    \centering
    \scriptsize
    \setlength{\tabcolsep}{2pt}
    \renewcommand{\arraystretch}{0.95}
    \begin{tabular*}{\columnwidth}{@{\extracolsep{\fill}}lcccc@{}}
    \toprule
    \textbf{Gait Feature} &
    \makecell{\textbf{Side}\\\textbf{(152)}} &
    \makecell{\textbf{Front}\\\textbf{(76)}} &
    \makecell{\textbf{Back}\\\textbf{(76)}} &
    \makecell{\textbf{Oblique}\\\textbf{(304)}} \\
    \midrule
    Cadence          & 0.74 & 0.89 & 0.89 & 0.83 \\
    Walking Speed    & 0.83 & 0.71 & 0.73 & 0.78 \\
    Step Length      & 0.40 & 0.30 & 0.31 & 0.35 \\
    Step Width       & 0.29 & 0.73 & 0.63 & 0.45 \\
    Stooped Posture  & 0.64 & 0.31 & 0.53 & 0.29 \\
    Arm Swing        & 0.67 & 0.46 & 0.86 & 0.78 \\
    \bottomrule
    \end{tabular*}
\end{table}
To further assess viewpoint sensitivity, we report both a view-averaged comparison across methods and a detailed view-wise breakdown for the HMR-based WHAM pipeline.
As shown in~\cref{tab:view-avg-comparison}, \model{} achieves the strongest overall view-averaged performance, with an average correlation of 0.79 compared with 0.65 for WHAM and 0.60 for PromptHMR. 
The gains are particularly clear for walking speed and step length,
while \model{} also remains competitive for step width. These
results suggest that direct RGB-to-gait estimation can be more
stable than pipelines that first reconstruct a human mesh for
several spatial gait quantities.
At the same time, HMR-based methods remain competitive for
pose-related quantities such as stooped posture and arm swing,
indicating that mesh reconstruction can still provide useful cues
for some gait features.

The detailed WHAM results in Tab.~\ref{tab:wham-view-analysis}
further illustrate the parameter-specific effect of viewpoint.
Walking speed remains relatively stable across camera directions,
whereas step width benefits substantially from frontal and back
views. Stooped posture is strongest from the side view, while arm
swing performs substantially better from the back and oblique
views than from the front. Step length remains comparatively
challenging across all viewpoints. These results support the
parameter-specific preferred-view protocol used in the main
benchmark and show that gait quantities derived through an HMR
pipeline can exhibit substantial viewpoint dependence.

\subsection{Additional Ablation Studies}
\begin{table}[t]
\caption{Effect of decoder design on gait-estimation performance. We report Pearson correlation ($r$). \textbf{Avg} denotes the Fisher $z$-transformed average.}
\label{tab:decoder-ablation}
\centering
\small
\setlength{\tabcolsep}{2pt}
\renewcommand{\arraystretch}{1.08}
\begin{tabular*}{0.98\columnwidth}{@{\extracolsep{\fill}}lccccccc@{}}
\toprule
\textbf{Decoder} 
& \textbf{Cad} 
& \textbf{Spd} 
& \textbf{Len} 
& \textbf{Wid} 
& \textbf{Stoop} 
& \textbf{Arm} 
& \textbf{Avg} \\
\midrule
AvgPool
& 0.92 & \textbf{0.88} & 0.66 & 0.65 & 0.74 & \textbf{0.91} & 0.82 \\
Transf. Dec.
& 0.93 & 0.87 & \textbf{0.68} & 0.64 & 0.70 & \textbf{0.91} & 0.82 \\
Cross-Attn.
& \textbf{0.94} & \textbf{0.88} & 0.67 & \textbf{0.67} & \textbf{0.75} & \textbf{0.91} & \textbf{0.84} \\
\bottomrule
\end{tabular*}
\end{table}
\paragraph{Choice of decoder.} \cref{tab:decoder-ablation} compares different decoder choices for mapping video encoder tokens to gait parameters. 
For the AvgPool baseline, we average-pool the encoder tokens into a single global representation and use separate linear heads to predict each gait parameter. 
The transformer decoder uses learnable gait queries with self-attention among queries followed by cross-attention to the encoder tokens, while cross-attention decoder removes query self-attention and directly attends from each gait query to the encoded video representation. 
Although all three designs achieve similar performance, the cross-attention decoder obtains the best Fisher-averaged correlation and slightly improves several gait parameters, including cadence, step width, and stooped posture. 
We therefore adopt the cross-attention decoder as the default design, as it provides a modest performance gain while keeping the decoder simple and task-specific.

\begin{figure}[!t]
    \centering
    \begin{subfigure}[t]{0.40\linewidth}
        \centering
        \includegraphics[
            width=\linewidth,
            height=0.40\textheight,
            keepaspectratio
        ]{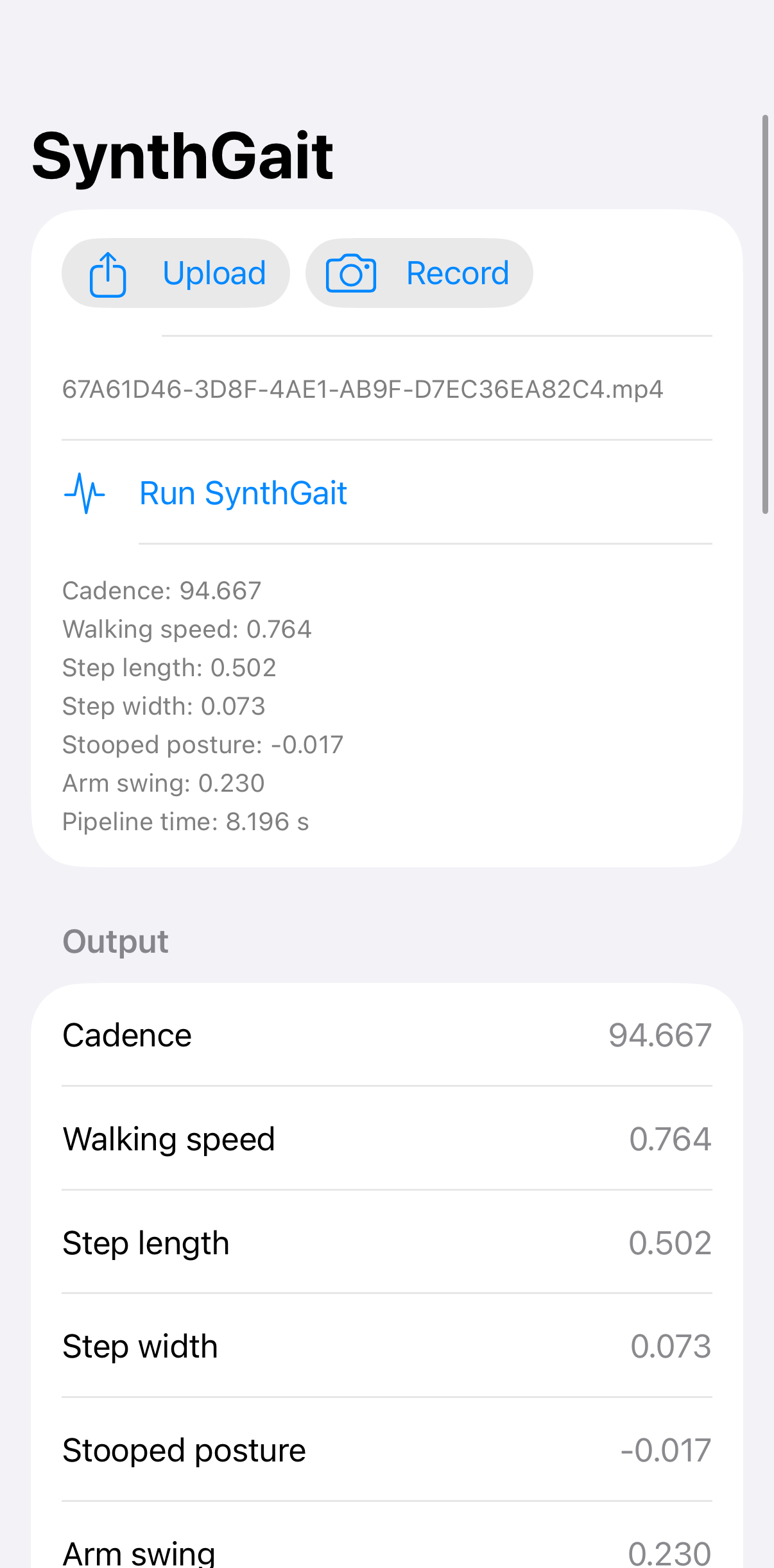}
        \caption{Prediction interface}
        \label{fig:mobile-app-output}
    \end{subfigure}
    \hspace{0.04\linewidth}
    \begin{subfigure}[t]{0.40\linewidth}
        \centering
        \includegraphics[
            width=\linewidth,
            height=0.40\textheight,
            keepaspectratio
        ]{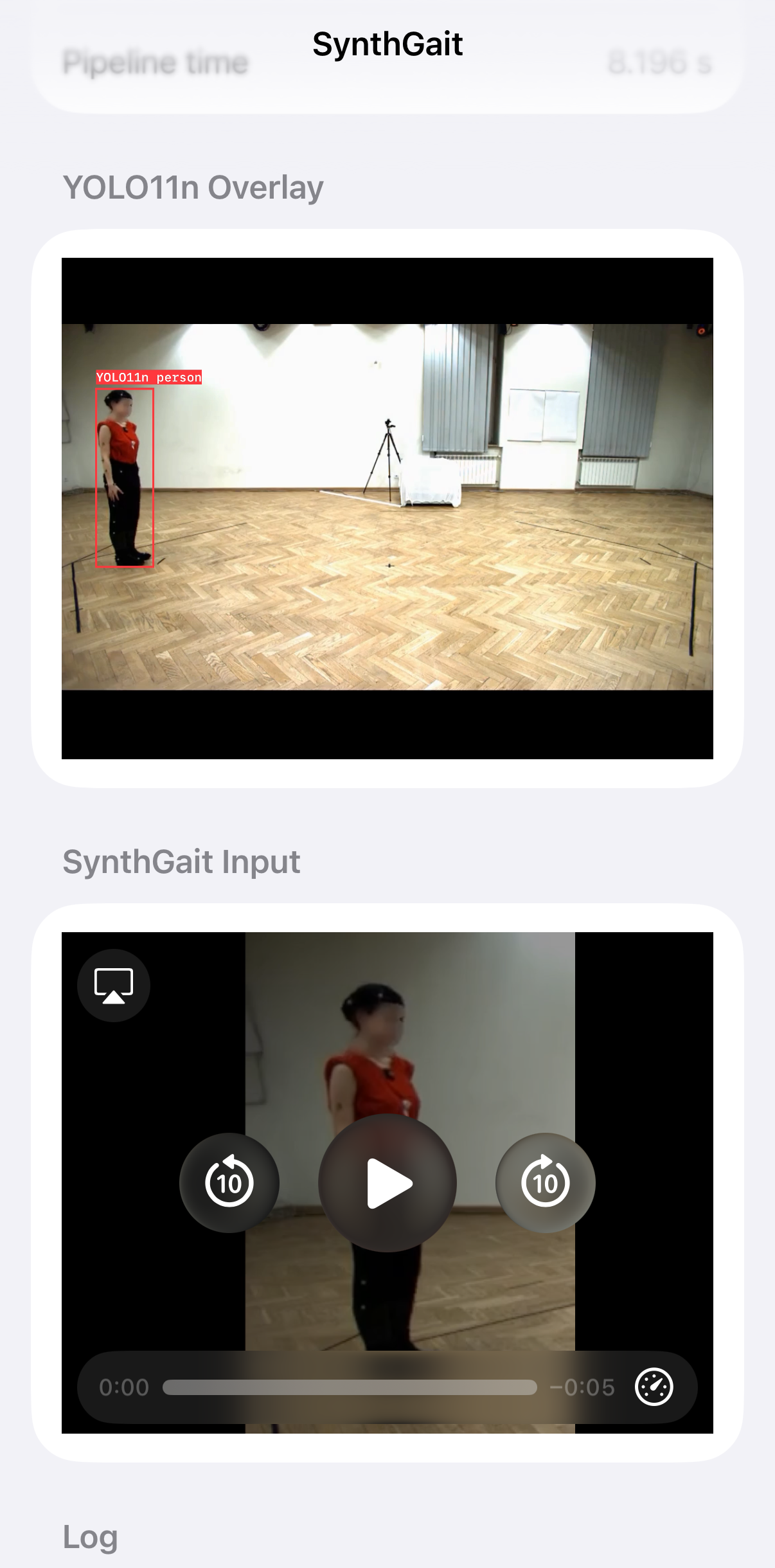}
        \caption{Tracking and input clip}
        \label{fig:mobile-app-tracking}
    \end{subfigure}

    \caption{Smartphone deployment of \model{}. The app tracks
    the walking subject with YOLO11n and runs \model{} on the
    cropped clip, completing the full on-device pipeline in 8.2 s.}
    \label{fig:mobile-app}
\end{figure}

\begin{table}[t]
\caption{Effect of encoder training strategy on gait-estimation performance. We report Pearson correlation ($r$); \textbf{Avg} is the Fisher $z$-transformed average.}
\label{tab:training-strategy-ablation}
\centering
\small
\setlength{\tabcolsep}{2pt}
\renewcommand{\arraystretch}{1.08}
\begin{tabular*}{0.98\columnwidth}{@{\extracolsep{\fill}}lccccccc@{}}
\toprule
\textbf{Strategy}
& \textbf{Cad}
& \textbf{Spd}
& \textbf{Len}
& \textbf{Wid}
& \textbf{Stoop}
& \textbf{Arm}
& \textbf{Avg} \\
\midrule
Frozen
& 0.75 & 0.87 & 0.68 & 0.62 & 0.64 & 0.88 & 0.76 \\
LoRA-16
& 0.88 & 0.88 & \textbf{0.72} & 0.56 & 0.68 & \textbf{0.91} & 0.80 \\
LoRA-32
& 0.88 & 0.87 & 0.56 & 0.53 & 0.73 & 0.90 & 0.81 \\
LoRA-64
& 0.86 & \textbf{0.90} & 0.69 & 0.53 & 0.72 & 0.89 & 0.80 \\
Full FT
& \textbf{0.94} & 0.88 & 0.67 & \textbf{0.67} & \textbf{0.75} & \textbf{0.91} & \textbf{0.84} \\
\bottomrule
\end{tabular*}
\end{table}
\paragraph{Effect of training strategy.}
\cref{tab:training-strategy-ablation} compares different strategies for adapting the V-JEPA2 video encoder to gait estimation. 
In the frozen setting, the encoder weights are kept fixed and only the decoder and regression heads are trained. 
For LoRA-based adaptation, we keep the pretrained encoder weights frozen and insert low-rank trainable adapters with ranks 16, 32, and 64. 
Finally, full fine-tuning updates the entire encoder together with the gait-query decoder and prediction heads. 
While LoRA improves over the frozen baseline and achieves competitive performance, full fine-tuning obtains the best Fisher-averaged correlation and improves several clinically relevant parameters, including cadence, step width, and stooped posture. 
We therefore use full fine-tuning as the default training strategy.

\begin{table}[t]
    \caption{
    Effect of input length on gait-estimation performance in terms of
    Pearson correlation ($r$). Avg denotes the Fisher $z$-transformed
    average across gait parameters.
    }
    \label{tab:input-length-ablation}
    \centering
    \scriptsize
    \renewcommand{\arraystretch}{1.05}
    \begin{tabular*}{\columnwidth}{@{\extracolsep{\fill}}lccccccc@{}}
    \toprule
    \textbf{Input}
    & \makecell{\textbf{Cad}}
    & \makecell{\textbf{W.}\\\textbf{Speed}}
    & \makecell{\textbf{Step}\\\textbf{Len}}
    & \makecell{\textbf{Step}\\\textbf{Wid}}
    & \makecell{\textbf{Stoop}\\\textbf{Post}}
    & \makecell{\textbf{Arm}\\\textbf{Swing}}
    & \makecell{\textbf{Avg}} \\
    \midrule
    $T{=}16$
    & 0.90 & \textbf{0.91} & \textbf{0.72} & 0.58 & 0.70 & 0.88 & 0.81 \\
    $T{=}32$
    & \textbf{0.94} & 0.88 & 0.67 & \textbf{0.67} & \textbf{0.75} & 0.91 & \textbf{0.84} \\
    $T{=}64$
    & 0.92 & 0.88 & 0.64 & 0.66 & 0.71 & \textbf{0.92} & 0.82 \\
    \bottomrule
    \end{tabular*}
\end{table}
\noindent\textbf{Input video length.}
We additionally study the effect of the number of input frames in
Tab.~\ref{tab:input-length-ablation}. Using $T=32$ frames achieves
the highest overall performance, with a Fisher $z$-averaged
correlation of $0.84$. Compared with $T=16$, the additional
temporal context improves several gait parameters, including
cadence, step width, and stooped posture. Increasing the input
length to $T=64$ does not provide consistent gains and slightly
reduces the overall correlation. We therefore use $T=32$ frames
for the final model.

\subsection{Controlled Occlusion Robustness}
\label{sec:occlusion-robustness}

\begin{table}[t]
    \centering
    \caption{
    Controlled occlusion robustness on GPJATK. We report Pearson
    correlation ($r$) and the Fisher $z$-transformed average.
    P denotes persistent occlusion over the full clip, and T denotes
    transient occlusion over one contiguous 50\% segment of the clip.
    $\Delta$ denotes the drop in average correlation relative to clean
    input. CIs are obtained from 10,000 paired participant-level
    bootstrap replicates.
    }
    \label{tab:occlusion}
    \scriptsize
    \setlength{\tabcolsep}{2.4pt}
    \resizebox{\columnwidth}{!}{
    \begin{tabular}{lcccccccc}
    \toprule
    Condition & Cad & Spd & Len & Wid & Stoop & Arm & Avg &
    $\Delta$ [95\% CI] \\
    \midrule
    Clean
    & .939 & .882 & .674 & .667 & .748 & .915
    & .837 & .000 \\
    
    Upper-P
    & .933 & .874 & .675 & .688 & .199 & .863
    & .775 & .062 [.038,.089] \\
    
    Upper-T
    & .930 & .875 & .652 & .668 & .732 & .886
    & .820 & .017 [.004,.033] \\
    
    Lower-P
    & .918 & .874 & .600 & .269 & .782 & .898
    & .789 & .049 [.031,.071] \\
    
    Lower-T
    & .926 & .856 & .569 & .600 & .762 & .898
    & .807 & .030 [.019,.047] \\
    
    Random-P
    & .925 & .863 & .578 & .487 & .572 & .781
    & .748 & .089 [.066,.123] \\
    
    Random-T
    & .930 & .851 & .600 & .651 & .731 & .884
    & .807 & .030 [.017,.052] \\
    \bottomrule
    \end{tabular}
    }
    \vspace{1mm}
\end{table}

\begin{figure}[t]
    \centering
    \includegraphics[width=\columnwidth]{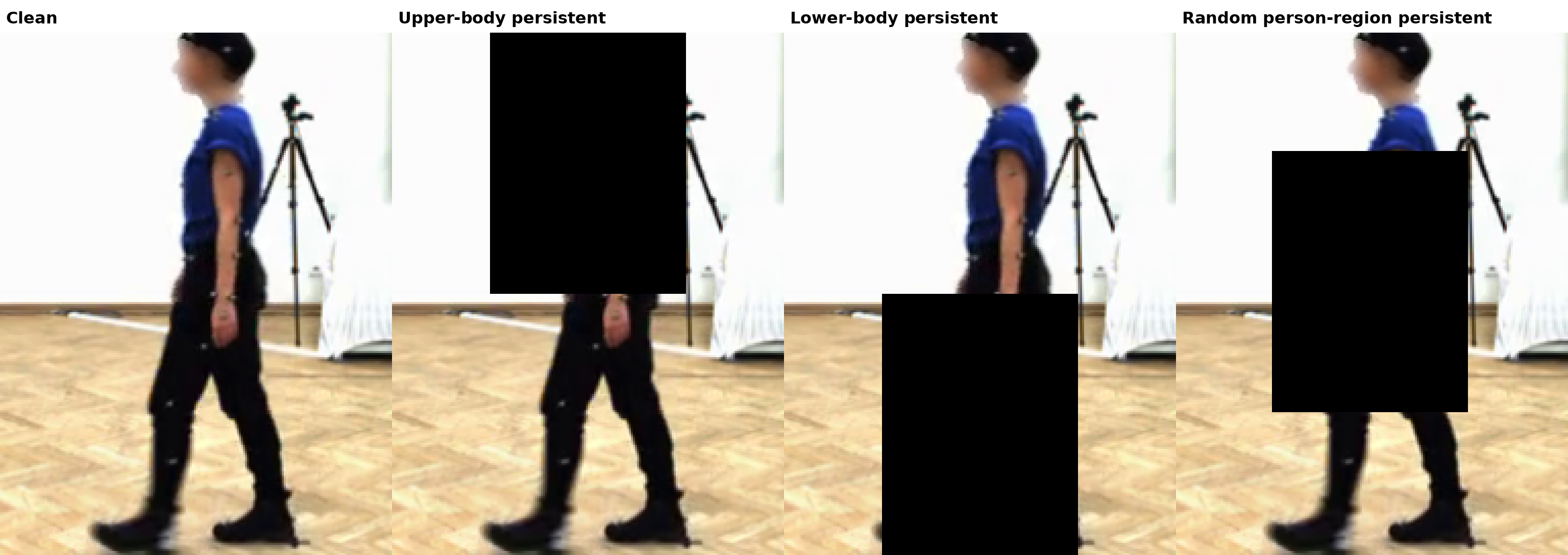}
    \caption{
    Controlled spatial occlusion settings. The same GPJATK frame
    is shown with clean input and persistent upper-body, lower-body,
    and random person-region occlusion. Each mask covers approximately
    25\% of the person-centered crop and remains fixed throughout
    the clip.
    }
    \label{fig:occlusion-spatial}
\end{figure}

\begin{figure}[t]
    \centering
    \includegraphics[width=\columnwidth]{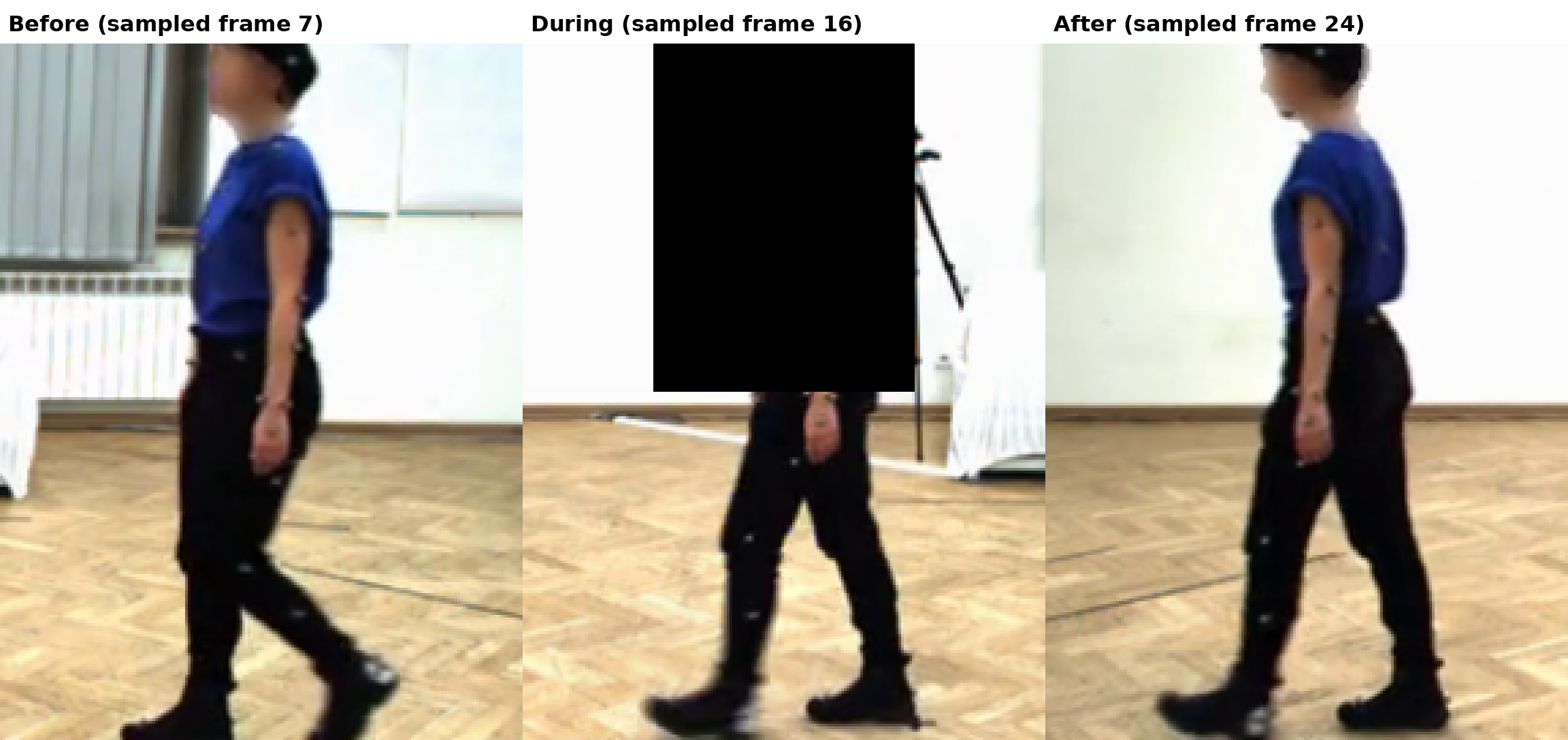}
    \caption{
    Transient occlusion setting. Example frames from the same
    GPJATK clip before, during, and after the occluded interval.
    The mask is applied over one contiguous 50\% segment of the
    sampled frames.
    }
    \label{fig:occlusion-transient}
\end{figure}

To evaluate robustness to incomplete visual evidence, we apply
controlled occlusions to GPJATK videos at inference time without
retraining \model{}. Each occlusion covers approximately 25\%
of the person-centered crop. We consider upper-body, lower-body,
and randomly positioned person-region masks, applied either
persistently throughout the full clip or transiently over one
contiguous 50\% temporal segment. The mask remains spatially
fixed within each clip, and all other preprocessing and evaluation
settings follow the preferred-view protocol used in the main paper.

As shown in Tab.~\ref{tab:occlusion}, the effect of occlusion is
strongly parameter-dependent. Persistent upper-body occlusion
primarily affects stooped posture, whose correlation decreases
from 0.75 to 0.20, while persistent lower-body occlusion most
strongly affects step width, which decreases from 0.67 to 0.27,
and also reduces step-length performance. Random persistent
occlusion produces the largest overall degradation, reducing the
Fisher $z$-averaged correlation from 0.84 to 0.75. In all three
spatial settings, transient occlusion produces a smaller overall
drop than persistent occlusion, indicating that \model{} can
partially recover when unoccluded temporal evidence remains
available. Paired participant-level bootstrap analysis further
shows that the Fisher-averaged degradation is positive for every
occlusion condition, with all 95\% confidence intervals excluding
zero. Figures~\ref{fig:occlusion-spatial} and
\ref{fig:occlusion-transient} illustrate the spatial and temporal masking protocols, respectively.

\subsection{Statistical Uncertainty of Headline Comparisons}
\label{sec:bootstrap-comparisons}

\begin{table}[t]
    \centering
    \caption{
    Paired participant-bootstrap comparison on GPJATK.
    Entries report $\Delta r=r_{\mathrm{GXF}}-r_{\mathrm{baseline}}$
    with 95\% percentile confidence intervals from 10,000 bootstrap
    replicates. Positive values favor GaitXFormer. Avg denotes the
    difference in Fisher-$z$-averaged correlation.
    }
    \label{tab:bootstrap-comparisons}
    \scriptsize
    \setlength{\tabcolsep}{3.0pt}
    \resizebox{\columnwidth}{!}{
    \begin{tabular}{lcc}
    \toprule
    Metric & GXF $-$ STT$^\dagger$ & GXF $-$ WHAM \\
    \midrule
    Cadence
    & $+.025\;[+.011,+.048]$
    & $+.094\;[+.058,+.152]$ \\
    
    Walking speed
    & $+.033\;[-.002,+.085]$
    & $+.056\;[+.014,+.100]$ \\
    
    Step length
    & $-.059\;[-.163,+.019]$
    & $+.272\;[+.047,+.454]$ \\
    
    Step width
    & $+.001\;[-.098,+.082]$
    & $-.019\;[-.145,+.108]$ \\
    
    Stoop posture
    & $+.044\;[-.053,+.172]$
    & $+.104\;[-.031,+.245]$ \\
    
    Arm swing
    & $-.007\;[-.030,+.025]$
    & $+.247\;[+.121,+.442]$ \\
    \midrule
    Avg
    & $+.013\;[-.008,+.035]$
    & $+.133\;[+.086,+.183]$ \\
    \bottomrule
    \end{tabular}
    }
\end{table}

We additionally quantify uncertainty in the principal GPJATK
comparisons using 10,000 paired participant-level bootstrap
replicates. Participants are resampled with replacement while all
sequences and synchronized views belonging to a participant are
kept together. For each gait parameter, methods are compared on
their common evaluated samples, while retaining the preferred-view
protocol used in the main paper.

Table~\ref{tab:bootstrap-comparisons} reports the signed difference
in Pearson correlation between \model{} and two representative
baselines. Relative to the task-specific STT$^\dagger$ model trained
on \dataset{}, \model{} obtains a slightly higher
Fisher-$z$-averaged correlation ($\Delta=0.013$), although the
95\% confidence interval includes zero, indicating comparable
overall performance under matched task supervision. In contrast,
\model{} shows a substantially larger average correlation than
WHAM ($\Delta=0.133$, 95\% CI $[0.086,0.183]$). At the individual
parameter level, the advantage over WHAM is particularly clear for
cadence, walking speed, step length, and arm swing.
\section{Mobile Deployment}
\label{sec:mobile-deployment}

To evaluate practical deployability, we implement a prototype \model{} mobile application that runs the full RGB-to-gait pipeline on a smartphone. As shown in Fig.~\ref{fig:mobile-app}, the app allows the user to upload or record a walking video, detects the walking subject using YOLO11n, extracts a person-centered input clip, and applies \model{} to estimate cadence, walking speed, step length, step width, stooped posture, and arm swing. The example in Fig.~\ref{fig:mobile-app}, tested on an iPhone 17, demonstrates that the complete pipeline, including detection, preprocessing, and \model{} inference, can be executed on-device in approximately 8 seconds. This result highlights that the proposed direct RGB-to-gait formulation is practical for lightweight mobile deployment.
\section{Ethics and Broader Impact}
\label{sec:ethics}

\dataset{} is intended to support research on video-based gait
analysis and evaluation rather than clinical diagnosis. Although the
estimated gait parameters are clinically relevant, predictions from
\model{} should not be interpreted as medical assessments without
appropriate validation in the target population and acquisition
setting. Performance may vary under factors such as severe occlusion,
assistive devices, clothing, camera placement, and population
characteristics that are not fully represented by the current
benchmarks.

Synthetic video provides a way to increase the diversity and scale of
gait data without collecting additional identifiable video from human
participants. At the same time, synthetic data do not eliminate biases
inherited from the underlying motion sources, rendering pipeline, or
evaluation datasets. We therefore view \dataset{} as a research
resource for developing and benchmarking gait-estimation methods, and
recommend evaluation on appropriately governed real-world cohorts
before deployment in health-related applications.

\end{document}